\documentclass[11pt, a4paper, copyright, frontis-thu]{frontis}
\uselogo{}
\fvlogodividerfalse        %
\fvtitlecentertrue         %

\usepackage{amsmath,amssymb,amsthm}
\usepackage{graphicx}
\usepackage{booktabs}
\usepackage{multirow}
\usepackage{array}
\usepackage{xcolor}
\usepackage{colortbl}
\usepackage{hyperref}
\usepackage{enumitem}
\usepackage{caption}
\usepackage{subcaption}
\usepackage{tabularx}
\usepackage{wrapfig}
\usepackage{makecell}
\usepackage{xspace}
\usepackage{tikz}
\usetikzlibrary{positioning,arrows.meta,shapes,fit,backgrounds}
\usepackage{fontawesome5}

\definecolor{badgebg}{HTML}{E5F6F8}
\definecolor{badgeborder}{HTML}{8DCFD6}

\newcommand{\badgeicon}[1]{\makebox[1.0em][c]{#1}}

\newcommand{\buildbadge}[2]{%
\tikz[baseline=(text.base)]{
  \node[
    fill=badgebg,
    draw=badgeborder,
    rounded corners=4pt,
    minimum height=20pt,
    inner xsep=6pt,
    inner ysep=2pt
  ] (text) {%
    {\fontfamily{ppl}\selectfont\small\strut
      \badgeicon{#1}\kern0.35em #2}%
  };
}%
}

\newsavebox{\badgeLB}
\savebox{\badgeLB}{%
\buildbadge{\textcolor{black}{\faGlobe}}{Leaderboard}%
}

\newsavebox{\badgeGH}
\savebox{\badgeGH}{%
\buildbadge{\textcolor{black}{\faGithub}}{GitHub}%
}

\newsavebox{\badgeHF}
\savebox{\badgeHF}{%
\buildbadge{%
  \raisebox{-0.05em}{\includegraphics[height=0.85em]{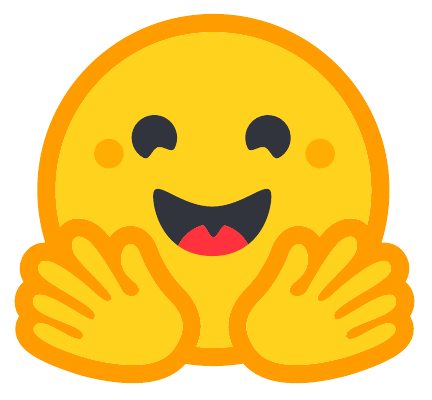}}%
}{HuggingFace}%
}

\hypersetup{
  colorlinks=true,
  linkcolor=blue!60!black,
  citecolor=blue!60!black,
  urlcolor=blue!60!black
}

\newcommand{\benchmark}{{\fontfamily{ppl}\selectfont NatureBench}\xspace}

\definecolor{nb1}{HTML}{EAF1FA}
\definecolor{nb2}{HTML}{CFE0F3}
\definecolor{nb3}{HTML}{A8C8E8}
\definecolor{nb4}{HTML}{7DAAD9}
\definecolor{nb5}{HTML}{5189C8}
\newcommand{\heatScolor}[1]{%
  \ifdim#1pt>0pt%
    \ifdim#1pt>27pt\cellcolor{nb5}%
    \else\ifdim#1pt>20pt\cellcolor{nb4}%
    \else\ifdim#1pt>12pt\cellcolor{nb3}%
    \else\ifdim#1pt>5pt\cellcolor{nb2}%
    \else\cellcolor{nb1}%
    \fi\fi\fi\fi%
  \fi}
\newcommand{\heatMcolor}[1]{%
  \ifdim#1pt>0pt%
    \ifdim#1pt>50pt\cellcolor{nb5}%
    \else\ifdim#1pt>38pt\cellcolor{nb4}%
    \else\ifdim#1pt>25pt\cellcolor{nb3}%
    \else\ifdim#1pt>12pt\cellcolor{nb2}%
    \else\cellcolor{nb1}%
    \fi\fi\fi\fi%
  \fi}
\newcommand{\hs}[1]{\heatScolor{#1}#1}
\newcommand{\hsb}[1]{\heatScolor{#1}\textbf{#1}}
\newcommand{\hsu}[1]{\heatScolor{#1}\underline{#1}}
\newcommand{\hm}[1]{\heatMcolor{#1}#1}
\newcommand{\hmb}[1]{\heatMcolor{#1}\textbf{#1}}
\newcommand{\hmu}[1]{\heatMcolor{#1}\underline{#1}}
\definecolor{ob1}{HTML}{FDF1E0}
\definecolor{ob2}{HTML}{FBDDB8}
\definecolor{ob3}{HTML}{F6C283}
\definecolor{ob4}{HTML}{EFA24E}
\definecolor{ob5}{HTML}{E07B2A}
\newcommand{\heatScolorD}[1]{%
  \ifdim#1pt>0pt%
    \ifdim#1pt>27pt\cellcolor{ob5}%
    \else\ifdim#1pt>20pt\cellcolor{ob4}%
    \else\ifdim#1pt>12pt\cellcolor{ob3}%
    \else\ifdim#1pt>5pt\cellcolor{ob2}%
    \else\cellcolor{ob1}%
    \fi\fi\fi\fi%
  \fi}
\newcommand{\heatMcolorD}[1]{%
  \ifdim#1pt>0pt%
    \ifdim#1pt>50pt\cellcolor{ob5}%
    \else\ifdim#1pt>38pt\cellcolor{ob4}%
    \else\ifdim#1pt>25pt\cellcolor{ob3}%
    \else\ifdim#1pt>12pt\cellcolor{ob2}%
    \else\cellcolor{ob1}%
    \fi\fi\fi\fi%
  \fi}
\newcommand{\ds}[1]{\heatScolorD{#1}#1}
\newcommand{\dm}[1]{\heatMcolorD{#1}#1}

\definecolor{gapneg}{HTML}{C8553D} %
\definecolor{gappos}{HTML}{2E7D5B} %
\definecolor{relfill}{HTML}{4C78A8} %
\newlength{\barlen}
\newcommand{\divbar}[1]{%
  \begin{tikzpicture}[baseline=-0.6ex]
    \draw[black!12,line width=0.2pt] (-5mm,-0.85mm) rectangle (5mm,0.85mm);%
    \draw[black!30,line width=0.3pt] (0,-1.1mm) -- (0,1.1mm);%
    \pgfmathsetlength{\barlen}{min(abs(#1)/0.42,1)*5mm}%
    \ifdim#1pt<0pt%
      \fill[gapneg] (0,-0.85mm) rectangle (-\barlen,0.85mm);%
    \else%
      \fill[gappos] (0,-0.85mm) rectangle (\barlen,0.85mm);%
    \fi%
  \end{tikzpicture}}
\newcommand{\fillbar}[1]{%
  \begin{tikzpicture}[baseline=-0.6ex]
    \draw[black!18,line width=0.2pt] (0,-0.85mm) rectangle (8mm,0.85mm);%
    \pgfmathsetlength{\barlen}{#1/100*8mm}%
    \fill[relfill] (0,-0.85mm) rectangle (\barlen,0.85mm);%
  \end{tikzpicture}}
\newcommand{\mg}[1]{$#1$\,\divbar{#1}}
\newcommand{\mgb}[1]{$\mathbf{#1}$\,\divbar{#1}}
\newcommand{\mgu}[1]{$\underline{#1}$\,\divbar{#1}}
\newcommand{\mn}[1]{{\footnotesize\textcolor{black!55}{$#1$}}}
\newcommand{\mnb}[1]{{\footnotesize\textcolor{black!55}{$\mathbf{#1}$}}}
\newcommand{\mnu}[1]{{\footnotesize\textcolor{black!55}{$\underline{#1}$}}}
\newcommand{\rel}[1]{#1\,\fillbar{#1}}
\newcommand{\relb}[1]{\textbf{#1}\,\fillbar{#1}}
\newcommand{\relu}[1]{\underline{#1}\,\fillbar{#1}}

\newcommand{\gimp}{g}                    %

\usepackage[sort&compress]{natbib}
\setheadertext{\benchmark}
\hypersetup{colorlinks=true, citecolor=darkblue, linkcolor=darkblue, urlcolor=darkblue}

\title{\benchmark: Can Coding Agents Match the Published SOTA of Nature-Family Papers?}

\author{%
  Horizon Research, Frontis.AI \quad Tsinghua University \\
  Correspondence: \texttt{zhangkaiyan@frontis.cn} \\\vspace{4pt}
  \rule{0pt}{1.5em}%
  \makebox[\linewidth][c]{%
    \href{https://frontisai.github.io/NatureBench/}{\usebox{\badgeLB}}\hspace{0.65em}%
    \href{https://github.com/FrontisAI/NatureBench}{\usebox{\badgeGH}}\hspace{0.65em}%
    \href{https://huggingface.co/datasets/FrontisAI/NatureBench}{\usebox{\badgeHF}}%
  }%
  \vspace{-6pt}%
}

\date{\today}
\makeatletter
\gdef\theabstract{
We introduce \textbf{NatureBench}, a cross-discipline benchmark of 90 tasks distilled from peer-reviewed Nature-family publications, designed to evaluate whether AI coding agents can move beyond \emph{reproduction} toward \emph{discovery} on real scientific problems.
NatureBench is built on \textbf{NatureGym}, an automated pipeline that constructs a standardized, per-task containerized environment from a source paper, addressing the environment-fragmentation problem that has limited the credibility of prior agent-on-research benchmarks.
Evaluating twelve frontier agent configurations under a strict web-search-disabled protocol, we find that the strongest model surpasses SOTA on only $17.8\%$ of tasks under the $g>0.1$ criterion.
Analysis of method pathways reveals that agents succeed primarily through methodological translation, converting scientific tasks into familiar supervised prediction problems, rather than through genuine scientific invention.
Failures are dominated by wrong method choice and insufficient compute budget, not by task misunderstanding.
We release the benchmark, the NatureGym pipeline, and a public leaderboard with maintainer-side reproduction.
}
\makeatother

\begin{document}

\maketitle
\vspace{-0.9em}
\begin{figure}[!ht]
\centering
\makebox[\linewidth][l]{%
  \includegraphics[width=\linewidth]{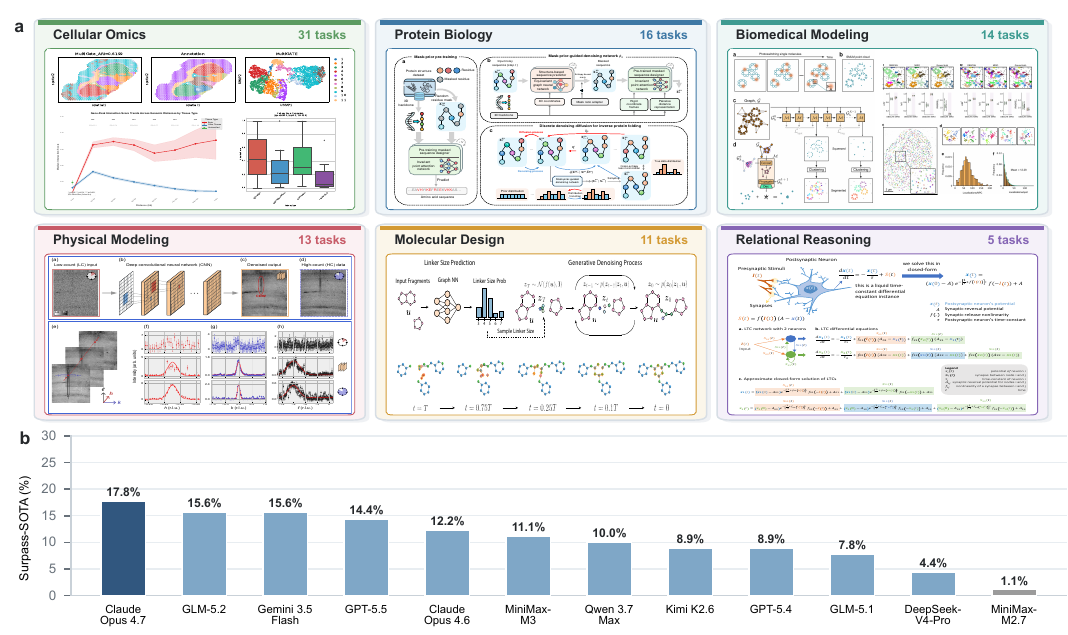}%
}
\captionsetup{font=small,skip=2pt}
\caption{\textbf{\benchmark overview.} (a) Six task domains with representative source figures \citep{miao2025multigate,bai2025mask,pineda2025enhanced,oppliger2024weak,igashov2024equivariant,hasani2022closed}. (b) NatureBench leaderboard by Surpass-SOTA ($g>0.1$) across twelve models.}
\label{fig:intro-main-results}
\end{figure}

\newpage
\begingroup
\setlength{\baselineskip}{0.9\baselineskip}   %
\tableofcontents
\endgroup
\newpage

\section{Introduction}
\label{sec:intro}

AI coding agents are rapidly moving toward autonomous scientific research~\citep{karpathy2026autoresearch,lu2026towards,gottweis2026accelerating}, from reproducing published implementations to conducting end-to-end research workflows.
As these systems begin to target real scientific problems, rigorous evaluation becomes critical: without reliable benchmarks, it is impossible to tell whether an agent is genuinely advancing the state of the art or merely fitting familiar patterns to new data.

However, existing benchmarks for evaluating agent capabilities on scientific research have several limitations.
Paper-based benchmarks~\citep{starace2025paperbench,siegel2024corebench,wang2026firebench} measure whether an agent can re-implement a published method, but stop short of the more consequential question: can an agent \emph{discover} a competitive method on its own?
Engineering-optimization benchmarks~\citep{shern2024mlebench,rank2026posttrainbench,nathani2025mlgym,qiang2026mle} target Kaggle competitions or post-training tasks, which do not require the domain reasoning, specialized tooling, or cross-discipline knowledge that characterize research in the natural sciences, and suffer from environment fragmentation that makes independent re-running fragile.
Credibly evaluating whether autonomous research agents can advance the frontier of AI-for-Science requires a benchmark that is both challenging and bidirectional. It must test \emph{discovery}, whether an agent can devise methods that surpass the published state of the art, on \emph{genuine scientific problems} drawn from the natural sciences rather than on engineering proxies.

We present \textbf{\benchmark}, a cross-discipline benchmark of 90 tasks distilled from peer-reviewed Nature-family publications, designed to evaluate whether AI coding agents can move beyond reproduction toward discovery.
\benchmark simultaneously extends both axes: the \textbf{PaperBench axis} from Understanding $\to$ Coding to Discovery, and the \textbf{PostTrainBench axis} from Engineering Optimization to Science.
It is built on \textbf{NatureGym}, an automated pipeline that converts a published paper into a containerized task package comprising a task brief, the paper's dataset, a held-out test set with hidden ground truth, and an automated evaluator, addressing the environment-fragmentation problem in prior benchmarks.
We collect approximately $5{,}500$ papers from ten Nature-family journals published between 2022 and 2025 and apply a three-stage build-then-verify pipeline to yield the final $90$ task packages (Figure~\ref{fig:gym-pipeline}).
An information firewall removes the source method from each package, so agents must discover solutions rather than reproduce them.
The benchmark spans six scientific task domains (cellular omics, protein biology, biomedical modeling, physical modeling, molecular design, and relational reasoning) and uses a SOTA-normalized relative gap $g$ as the primary metric, supplemented by a post-hoc validity judge that detects shortcut behaviors such as output fabrication and feedback gaming.

We evaluate twelve agents spanning three coding-agent harnesses (Claude Code, Codex CLI, Gemini CLI) and twelve frontier models under a strict web-search-disabled protocol as shown in Figure~\ref{fig:intro-main-results}.
The strongest agent, Claude Opus 4.7, surpasses the published SOTA ($g > 0.1$) on only $17.8\%$ of tasks and matches it on $47.8\%$.
A ten-agent behavioral analysis over $900$ task--agent runs reveals that success is driven primarily by methodological translation, where agents convert scientific tasks into familiar supervised-prediction problems, accounting for $45.5\%$ of validated successes, rather than by scientific invention.
Failures are dominated by wrong method choice ($45.1\%$) and insufficient compute budget ($24.4\%$), not by task misunderstanding.
Our contributions are as follows:
\begin{itemize}[leftmargin=1em, itemsep=2pt, topsep=4pt]
    \item \textbf{NatureGym}, an automated pipeline that constructs reproducible, containerized per-task environments from Nature-family papers, addressing the environment-fragmentation problem that has limited the credibility of prior agent-on-research benchmarks.
    \item \textbf{NatureBench}, a benchmark of $90$ Nature-sourced tasks across six scientific task domains with a Discovery-oriented evaluation protocol (Surpass-SOTA, Match-SOTA, validity judge) that separates genuine algorithmic progress from engineering optimization and shortcut-taking.
\end{itemize}

\section{NatureGym}
\label{sec:naturegym}
We introduce \textbf{NatureGym}, a pipeline that turns a published Nature-family paper into a ready-to-run agentic task.
Each task is a containerized package comprising a task brief, the dataset, a held-out test set, an automated evaluator, and a SOTA anchor score.
NatureGym standardizes papers with heterogeneous formats, toolchains, and data modalities into one reproducible task format, while imposing an information firewall that withholds the original method so that agents must discover solutions rather than reproduce them.

\subsection{Pipeline Overview}
\label{sec:gym:overview}

\begin{figure}[t]
    \centering
    \includegraphics[width=1.0\linewidth]{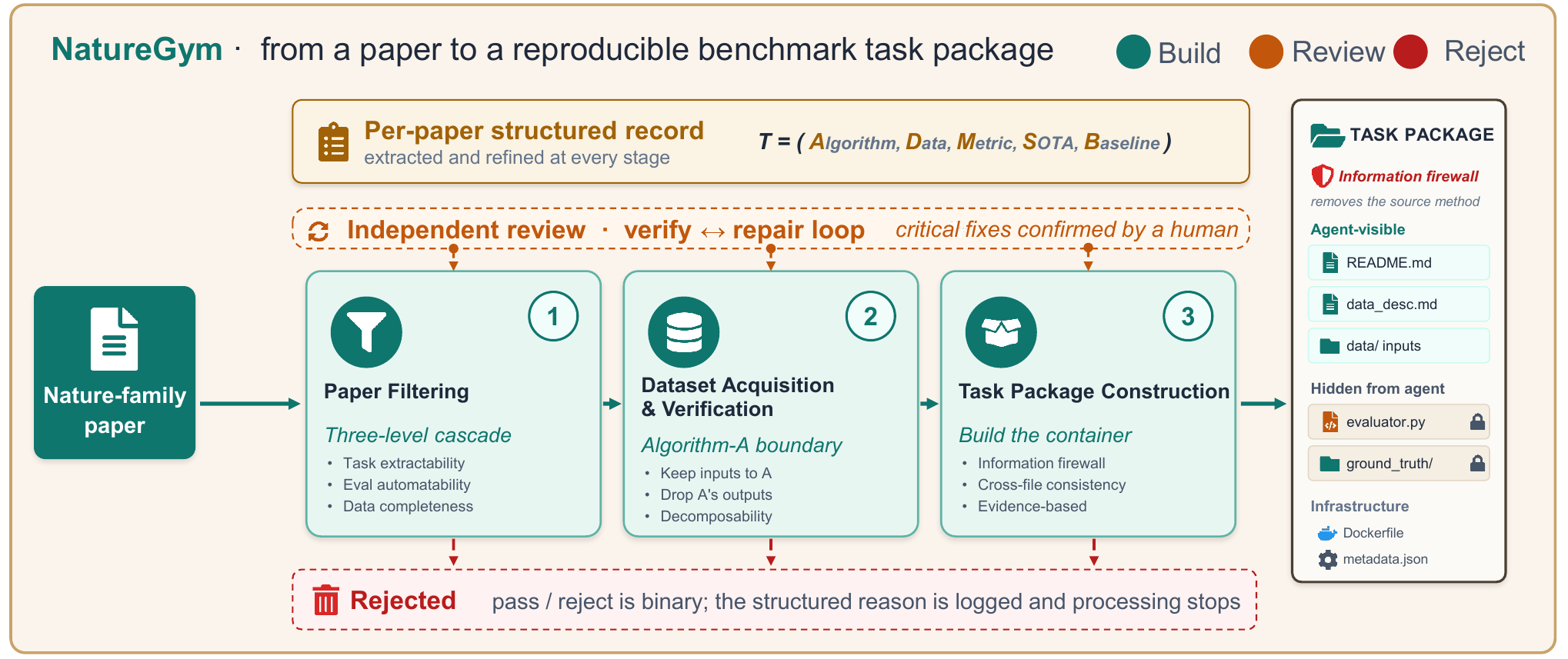}
    \caption{
    \textbf{The NatureGym pipeline.} Three review-gated stages turn one Nature-family paper into a containerized task package, refining a shared per-paper record $T=(A,D,M,S,B)$ along the way. An information firewall removes the source method, so the agent receives only dataset inputs, task brief, and a held-out test set, and tries to discover rather than reproduce.
    }
    \label{fig:gym-pipeline}
\end{figure}

As shown in Fig.~\ref{fig:gym-pipeline}, NatureGym builds each task through three stages: Paper Filtering (\S\ref{sec:gym:filter}), Dataset Acquisition and Verification (\S\ref{sec:gym:data}), and Task Package Construction (\S\ref{sec:gym:build}).
Each stage ends with an independent review that catches and corrects errors through a verify--repair loop before the next stage begins.

Every stage serves two purposes. First, it makes a binary pass-or-reject decision, terminating all downstream processing for rejected papers. Second, it extracts and refines structured task information into a per-paper record that accumulates across stages, so that task package construction can consume this record directly without re-reading the paper.

We represent each task as a tuple $T=(A,D,M,S,B)$, namely a core algorithm $A$, a dataset $D$, a metric $M$, a SOTA score $S$, and an optional baseline $B$.
The pipeline starts to fill in this tuple at the filtering stage and refines it in every later stage.
Every stage is run by an LLM agent, and a human confirms the critical corrections that each review surfaces.

\subsection{Paper Filtering}
\label{sec:gym:filter}

Paper filtering identifies candidate papers suitable for task construction through three steps: preprocessing, a three-level cascade filter, and an adversarial review.

\paragraph{Preprocessing.}

Each paper is converted into three structured components that the subsequent filtering stages consume.
After retaining only research articles and dropping non-research content (e.g., news, editorials, corrections, reviews), we produce from each article: (i) markdown text preserving document structure and formulas with citation markers removed;
(ii) full-page screenshots of every figure and table;
and (iii) a section-tagged list of hyperlinks categorized as data, code, supplementary material, or other, with surrounding context.

\paragraph{Three-level filtering.}
We then apply three filtering levels, each targeting a distinct feasibility dimension: task extractability, evaluation automatability, and data completeness.

\begin{itemize}[topsep=0pt, partopsep=0pt, leftmargin=12pt, itemsep=5pt]
\item \textbf{Level~1: Task.} The paper's core contribution must yield an extractable ML task: an algorithmic innovation, an ML formulation of a scientific problem, or a domain adaptation of an established method. We exclude papers in which ML serves only as an auxiliary tool, non-computational studies (wet-lab experiments, pure theory, hardware), and tasks that require physical interaction.

\item \textbf{Level~2: Evaluation.} The paper must claim state-of-the-art performance on a quality-related metric, rather than on speed, cost, or interpretability. Moreover, this metric must admit a deterministic, fully automated evaluation that does not rely on human judgment, external service dependencies, or components of the algorithm itself.

\item \textbf{Level~3: Data.} All data must match the version used in the paper and be publicly accessible without application or authentication. The dataset must be complete, with a development set $D_\text{dev}$ and an evaluation set $D_\text{eval}$ that further decomposes into test inputs $X_\text{test}$ and reference answers $Y_\text{ref}$. At least one evaluation instance must satisfy all conditions. We further tag each dataset by volume (Tier~S $<$\,1\,GB, Tier~M\,1--50\,GB, Tier~L\,$>$\,50\,GB) and reject papers whose data exceeds 50\,GB.
\end{itemize}

\paragraph{Filtering review.}
Before entering the costly data-acquisition stage, a separate adversarial pass re-examines every paper that passed, targeting false positives.
It rechecks both the pass-or-reject decision and the extracted task information, writing corrections back into the per-paper record. Critical overrides are confirmed by a human.

\subsection{Dataset Acquisition and Verification}
\label{sec:gym:data}

Papers that pass filtering enter dataset acquisition, where we download the data, determine the boundary separating the task definition from the paper's core algorithm, and re-verify data completeness against the actual files rather than the metadata-level probes of the filtering stage.

\paragraph{Dataset acquisition.}
We clone the linked code and data repositories and download the datasets by size tier and priority, taking the evaluation instances behind the paper's main results first.
Tier~S datasets are downloaded in full, while Tier~M datasets are downloaded one instance at a time under a cumulative size cap, and we skip the remaining instances once the cap is reached. Tier~L papers have already been
removed during filtering.

\paragraph{File-level firewall.}
To keep the information firewall intact, the agent must start exactly where the core algorithm $A$ starts, so it receives the inputs to $A$ but none of $A$'s operations or outputs.
We decide which files to keep by one question: \emph{is this file needed to define the task no matter which method is used?}
Files that define the task and are shared across methods are retained, including raw inputs that precede $A$, shared outputs of method-agnostic data preparation, and external resources.
Files that are specific to $A$ or produced by $A$ are excluded, including $A$'s own preprocessing, its intermediate or final outputs, and any irrelevant files.
We make each decision by reading the paper, the code, and the materialized data together.

\paragraph{Dataset verification and review.}
The filter judges feasibility from metadata alone, so we now re-run checks on the downloaded files. Two properties matter most.
\emph{Decomposability}: whether $D_\text{dev}$ separates from $D_\text{eval}$ using only sample-level splits and method-agnostic preparation (no algorithm or evaluation-time operations), and whether $X_\text{test}$ separates from $Y_\text{ref}$ while preserving all available features. We rate each split’s difficulty and reject infeasible cases. At this stage we only record the required split procedure. The actual partitioning is performed in \S\ref{sec:gym:build}.
\emph{Instance validity}: whether the retained evaluation instances correspond to a single research objective and include the core experiment. Non-core or analysis-only instances are discarded.
The check succeeds as long as at least one instance is complete.
A separate read-only review then cross-references the paper, code, and files to re-verify the $A$-boundary and all recorded descriptions. A fix step then repairs the record and reconciles the directory by removing surplus or leaking files and re-acquiring missing components, so that both the record and the data are ready for task construction. Cases with extensive corrections are confirmed by manual review.

\subsection{Task Package Construction}
\label{sec:gym:build}

\begin{table}[t]
\centering
\caption{\textbf{Task package structure produced by NatureGym.} Components under \texttt{problem/} are agent-visible; those under \texttt{evaluation/} are hidden from the agent.}
\label{tab:taskpkg}
\small
\begin{tabularx}{\textwidth}{llX}
\toprule
\textbf{Visibility} & \textbf{Component} & \textbf{Contents} \\
\midrule
\multirow{3}{*}{Agent-visible}
  & \texttt{problem/README.md}             & Task definition, evaluation metrics, output format, submission specification \\
  & \texttt{problem/data\_description.md}  & Dataset overview, file formats and schemas \\
  & \texttt{problem/data/}                 & Per-instance inputs (\emph{ground truth excluded}) \\
\midrule
\multirow{2}{*}{Hidden}
  & \texttt{evaluation/evaluator.py}       & Deterministic scoring function with input validation \\
  & \texttt{evaluation/ground\_truth/}     & Per-instance reference answers \\
\midrule
\multirow{2}{*}{Infrastructure}
  & \texttt{environment/Dockerfile}        & Per-task overlay on the shared base image \\
  & \texttt{metadata.json}                 & Domain, compute requirements, per-instance SOTA scores \\
\bottomrule
\end{tabularx}
\end{table}

Each paper that passes filtering and data verification is assembled into the task package layout of Table~\ref{tab:taskpkg}.
Construction and subsequent verification follow three principles:
(i) \textbf{Evidence-grounded fidelity}: every component and performance anchor must be supported by verified records and source evidence.
(ii) \textbf{Information firewall}: no file may reveal the source paper's identity or method, and task inputs must be separated from hidden references and scoring logic.
(iii) \textbf{Executable integrity}: all components must be mutually consistent in semantics and interfaces, and the package as a whole must pass both static checks and end-to-end execution.

\paragraph{Data organization.}
Following the decomposition procedure from \S\ref{sec:gym:data}, we route inputs to the agent-visible \texttt{problem/data/} and reference answers to the hidden \texttt{evaluation/ground\_truth/}, with the routing rule determined by the reference-answer type (static label, oracle function, or distributional statistic).
Instances whose required evaluation components cannot be sourced from public libraries or reimplemented from author code are excluded. Construction continues as long as at least one instance remains viable.

\paragraph{Task documentation.}
Each package ships two documents under the information-firewall constraint. \texttt{data\_description.md} is a technical reference for the files in \texttt{problem/data/}, covering dataset overview, formats, and schemas. \texttt{README.md} defines the task, evaluation metrics, output format, and submission specification, retaining only the quality metrics the paper uses for ranking and designating one primary metric per instance for aggregate scoring.
\texttt{metadata.json} records the scientific domain, compute requirements, and per-instance SOTA scores extracted from the paper text, tables, or figures.

\paragraph{Automated evaluator.}
The evaluator independently scores agent outputs, dispatching on the reference-answer type: it compares against the ground truth for \emph{Label} tasks, runs the scoring function for \emph{Oracle} tasks, and computes distributional statistics for \emph{Distribution} tasks. It validates output format and shape before scoring, and scores multi-instance tasks with failures isolated so that one does not affect the rest. We check the evaluator at build time with logic tests, smoke tests, comparison against author code where available, and verification of evaluator scores against the paper's reported values using the authors' released outputs.

\paragraph{Execution environment.}
A shared base image pre-installs core scientific and ML libraries. Task-specific dependencies are layered on top via per-task Dockerfiles, with a standalone build reserved for irreconcilable conflicts such as a different CUDA or Python version.

\paragraph{Package and environment review.}
Unlike the one-shot reviews of the previous stages, this review runs an iterative verify--repair loop.
A build-time self-audit first rechecks the task definition, SOTA scores, and firewall against the source paper.
Then 36 automated checks cover artifact completeness, cross-component consistency, the information firewall, benchmark-design conformance, and end-to-end dynamic testing. The last category runs a baseline solver through the full evaluation pipeline together with correctness and robustness probes.
Finally, the Docker image is built on a physical machine and smoke-tested for library availability and version correctness.
Failed checks trigger minimal targeted repairs and immediate re-verification. Issues that resist automated repair are escalated to human review.
The full check inventory and repair strategy are described in Appendix~\ref{app:package-review}.

\section{NatureBench}
\label{sec:naturebench}

\providecommand{\cmark}{\textcolor{green!45!black}{\(\checkmark\)}}
\providecommand{\xmark}{\textcolor{red!70!black}{\(\times\)}}

\definecolor{catblue}{HTML}{EFF5F9}       %
\definecolor{catbluestrong}{HTML}{D6E6F0}  %
\definecolor{catgreen}{HTML}{F1F7F1}       %
\definecolor{catgreenstrong}{HTML}{DCEEDD}  %
\definecolor{catorange}{HTML}{FBF6EA}      %
\definecolor{catorangestrong}{HTML}{F5ECCE} %
\definecolor{catpurple}{HTML}{F5F1F9}      %
\definecolor{catpurplestrong}{HTML}{E8DFF2} %

\begin{table}[t]
\centering
\caption{\textbf{Comparison with representative agent benchmarks.} \textbf{\# Tasks} reports the source-stated number of primary evaluation units. \textbf{Paper} indicates whether tasks are derived from source papers; \textbf{Science} indicates whether they address scientific domains beyond AI/ML methodology itself; and \textbf{Optimization} indicates whether agents maximize task performance rather than recover or assess known results. NatureBench uniquely combines paper-sourced tasks, scientific-domain coverage, and discovery-oriented evaluation.}
\label{tab:benchmark-comparison}
\footnotesize
\setlength{\tabcolsep}{3.2pt}
\renewcommand{\arraystretch}{1.07}
\resizebox{\textwidth}{!}{%
\begin{tabular}{@{}l l c c c c l l@{}}
\toprule
\textbf{Benchmark} & \textbf{Source} & \textbf{\# Tasks} & \textbf{Paper} & \textbf{Science} & \textbf{Optimization} & \textbf{Objective} & \textbf{Scoring anchor} \\
\midrule
\rowcolor{catbluestrong}
\multicolumn{8}{l}{\textit{ML Paper Replication}} \\
\rowcolor{catblue}
PaperBench~\citep{starace2025paperbench} & ICML papers & 20 & \cmark & \xmark & \xmark & paper replication & author rubrics \\
\rowcolor{catblue}
AutoExperiment~\citep{kim2025autoexperiment} & ML papers & 85 & \cmark & \xmark & \xmark & masked-code reproduction & gold outputs \\
\rowcolor{catblue}
FIRE-Bench~\citep{wang2026firebench} & LLM analysis papers & 30 & \cmark & \xmark & \xmark & finding rediscovery & paper claims \\
\midrule
\rowcolor{catgreenstrong}
\multicolumn{8}{l}{\textit{Scientific Paper Reproduction}} \\
\rowcolor{catgreen}
CORE-Bench~\citep{siegel2024corebench} & Code Ocean capsules & 270 & \cmark & \cmark & \xmark & result reproduction & manual outputs \\
\rowcolor{catgreen}
REPRO-Bench~\citep{hu2025reprobench} & social-science papers & 112 & \cmark & \cmark & \xmark & reproducibility assessment & expert labels \\
\rowcolor{catgreen}
ReplicationBench~\citep{ye2025replicationbench} & astrophysics papers & 111 & \cmark & \cmark & \xmark & result replication & reported values \\
\rowcolor{catgreen}
AutoMat~\citep{huang2026automat} & materials-science papers & 85 & \cmark & \cmark & \xmark & claim reproduction & expert annotations \\
\rowcolor{catgreen}
Collider-Bench~\citep{faroughy2026colliderbench} & LHC papers & 10 & \cmark & \cmark & \xmark & analysis reproduction & event yields \\
\midrule
\rowcolor{catorangestrong}
\multicolumn{8}{l}{\textit{Task-Performance Optimization}} \\
\rowcolor{catorange}
MLE-bench~\citep{shern2024mlebench} & Kaggle competitions & 75 & \xmark & \xmark & \cmark & ML engineering & Kaggle leaderboard \\
\rowcolor{catorange}
PostTrainBench~\citep{rank2026posttrainbench} & model--benchmark pairs & 28 & \xmark & \xmark & \cmark & LLM post-training & official instruct models \\
\rowcolor{catorange}
MLS-Bench~\citep{lyu2026mlsbench} & ML research problems & 140 & \xmark & \xmark & \cmark & method invention & human baselines \\
\rowcolor{catorange}
AutoLab~\citep{xu2026autolab} & expert-curated problems & 36 & \xmark & \xmark & \cmark & long-horizon optimization & baseline/human metrics \\
\midrule
\rowcolor{catpurplestrong}
\textbf{NatureBench (ours)} & \textbf{Nature-family papers} & \textbf{90} & \textbf{\cmark} & \textbf{\cmark} & \textbf{\cmark} & \textbf{method development} & \textbf{published SOTA} \\
\bottomrule
\end{tabular}%
}
\end{table}

In this section, we introduce \textbf{NatureBench}, a benchmark of 90 task packages spanning six scientific task domains, produced by applying NatureGym (\S\ref{sec:naturegym}) to Nature-family journal papers.
We describe the source corpus and pipeline funnel (\S\ref{sec:nb:corpus}), evaluation-time quality calibration (\S\ref{sec:nb:calib}), benchmark composition (\S\ref{sec:nb:stats}), and evaluation protocol (\S\ref{sec:nb:eval}).

Table~\ref{tab:benchmark-comparison} positions NatureBench relative to representative agent benchmarks. Existing work either grounds tasks in papers but targets reproduction rather than optimization (PaperBench~\citep{starace2025paperbench}, CORE-Bench~\citep{siegel2024corebench}, ReplicationBench~\citep{ye2025replicationbench}), or optimizes task performance but draws from Kaggle or ML-engineering problems rather than scientific papers (MLE-bench~\citep{shern2024mlebench}, PostTrainBench~\citep{rank2026posttrainbench}). NatureBench is the first to combine paper-sourced tasks, genuine scientific problems, and optimization-oriented evaluation scored against the published SOTA.

\subsection{Source Corpus}
\label{sec:nb:corpus}

We first bound the source pool with a journal-level selection policy, then run the NatureGym pipeline (\S\ref{sec:naturegym}) to progressively narrow the crawled candidates into a construction-ready set that enters calibration.

\paragraph{Journal selection.} We select source journals by three criteria.
First, accepted papers must contain concrete algorithmic contributions with numerical SOTA claims, providing a clear competition target for each task.
Second, the journal must include papers with available data, so that the underlying datasets are publicly recoverable without per-item manual approval.
Third, the journal's topical scope must cover scientific machine learning, the domain where automated-agent capability is least studied.
Accordingly, we select ten Nature-family journals: \emph{Nature Machine Intelligence}, \emph{Nature Communications}, \emph{Nature Methods}, \emph{Nature Materials}, \emph{Nature Biomedical Engineering}, \emph{Nature Energy}, \emph{Nature Biotechnology}, \emph{Nature Computational Science}, \emph{Nature Genetics}, and \emph{Nature Neuroscience}. The publication window is 2022--2025, chosen to balance corpus size against software-stack currency and data-contamination risk.
The final 90-task set draws from six of these journals. The other four retain no tasks after filtering, data verification, task construction, and calibration.

\paragraph{Pipeline funnel.} All collected papers pass through five phases: the three NatureGym stages (\S\ref{sec:naturegym}) bookended by an initial collection crawl and a final calibration step.
\emph{Collection} crawls ${\sim}$5{,}500 initial candidates from ten Nature-family journals.
\emph{Filtering} retains ${\sim}$2{,}500 research articles via an article-type filter, then applies three-stage filtering (\S\ref{sec:gym:filter}) to yield ${\sim}$200 papers.
\emph{Acquisition} acquires and verifies datasets (\S\ref{sec:gym:data}), narrowing to ${\sim}$180.
\emph{Construction} builds and verifies task packages (\S\ref{sec:gym:build}), retaining ${\sim}$160.
\emph{Calibration} removes defective tasks via evaluation-time quality calibration (\S\ref{sec:nb:calib}), finalizing the benchmark at 90~task packages.
Table~\ref{tab:funnel} reports counts at each step.

\begin{figure}[t]
\centering
\includegraphics[width=\linewidth]{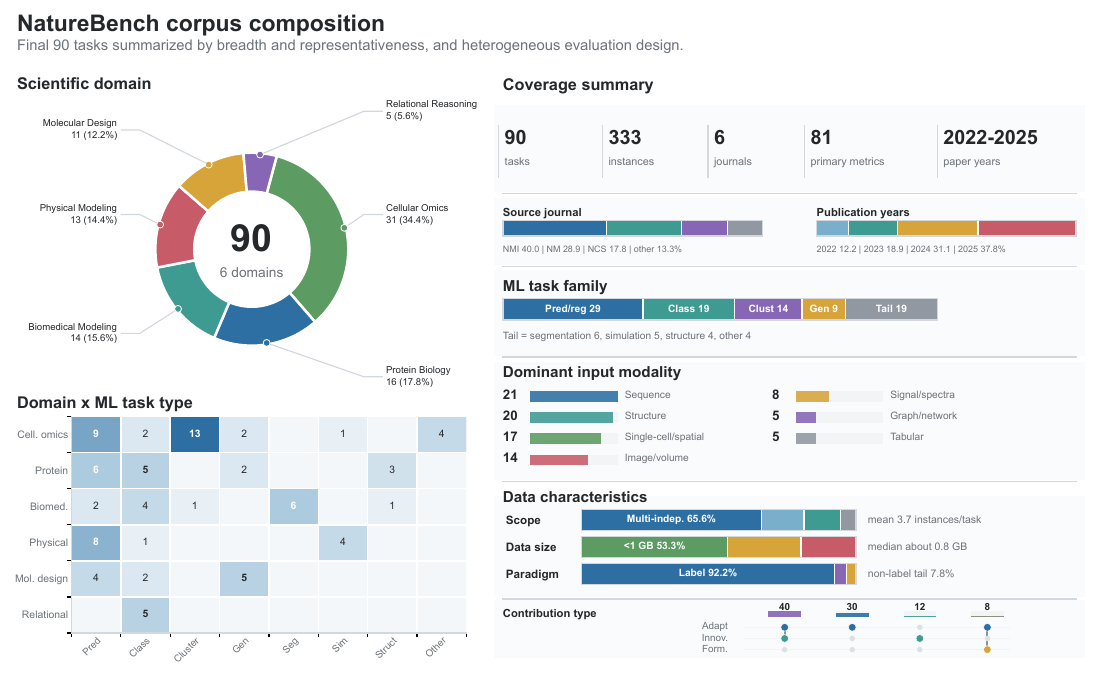}
\caption{\textbf{NatureBench coverage.}
Across 90 tasks, NatureBench spans six scientific domains and diverse ML task families while varying substantially in data modality, data characteristics, and source-paper contribution type.
}
\label{fig:nb:benchmark-statistics}
\end{figure}

\begin{table}[!ht]
\centering
\caption{\textbf{NatureGym pipeline funnel,} grouped into five phases aligned with the pipeline of \S\ref{sec:naturegym}. Counts marked with ``$\sim$'' are approximate, and only the final corpus size is exact.}
\label{tab:funnel}
\small
\begin{tabular}{llr}
\toprule
\textbf{Stage} & \textbf{Step} & \textbf{Papers retained} \\
\midrule
Collection    & Initial crawl from 10 Nature-family journals                 & $\sim$5{,}500 \\
Filtering     & Article-type filter (exclude non-research)                   & $\sim$2{,}500 \\
Filtering     & Three-level filtering~(\S\ref{sec:gym:filter})               & $\sim$200     \\
Acquisition   & Dataset acquisition and verification~(\S\ref{sec:gym:data})    & $\sim$180     \\
Construction  & Task construction~(\S\ref{sec:gym:build})     & $\sim$160     \\
Calibration   & Evaluation-time quality calibration~(\S\ref{sec:nb:calib})   & $\mathbf{90}$ \\
\bottomrule
\end{tabular}
\end{table}

\subsection{Benchmark Quality Calibration}
\label{sec:nb:calib}

Build-time verification (\S\ref{sec:naturegym}) guarantees only that a task package is structurally well-formed and runnable. Some defects surface only when an agent actually attempts to solve the task.
We therefore add an evaluation-time quality calibration before the main experiments, proceeding in three steps. Appendix~\ref{app:calibration-details} provides full details.

\paragraph{First-round diagnosis and repair.} We run Claude Opus 4.6 over all tasks in base mode and diagnose each case by combining the score, the agent trajectory, and the task package. Exposed defects include ground-truth leakage, distorted task definitions, metrics that fail to distinguish shortcuts from genuine solutions, evaluator inconsistencies, pipeline or environment errors, and missing data. Locally verifiable defects receive minimal repairs. Tasks with irreparable issues are dropped. Legitimate low scores are retained.

\paragraph{Reproduction-mode package audit.}
In reproduce mode, the agent additionally receives the source paper and is instructed to faithfully reproduce its method. We run Claude Opus 4.6 and DeepSeek-V4-Pro in this mode to audit whether each package genuinely supports the paper's approach, checking task description and data, evaluator, metadata anchors, and cross-component consistency.
After human review, 45 tasks are dropped for systematic defects and 17 receive minor repairs. The benchmark is finalized at 90 task packages.

\begin{figure}[t]
\centering
\includegraphics[width=\textwidth]{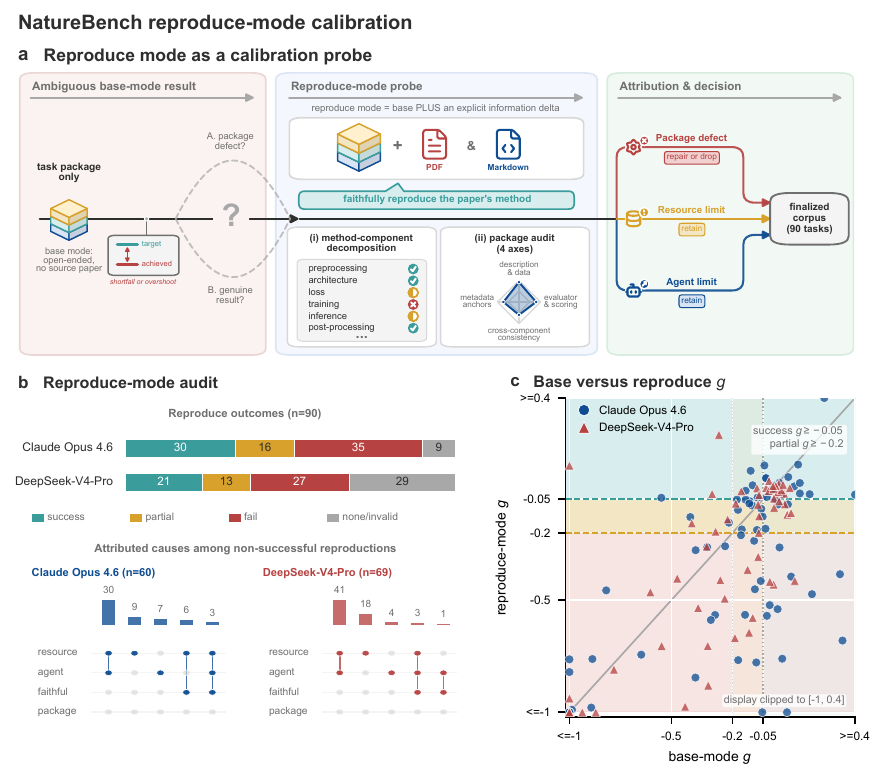}
\caption{\textbf{Calibrating the NatureBench corpus.}
\textbf{a}, The reproduce-mode calibration probe, from an ambiguous result to a clear attribution.
\textbf{b}, Reproduce-mode outcomes and attributed causes for Claude Opus 4.6 and DeepSeek-V4-Pro.
\textbf{c}, Per-task $\gimp$ in base versus reproduce mode.}
\label{fig:nb:benchmark-calibration}
\end{figure}

\paragraph{Reproducibility of the final set.}
On the finalized 90 tasks, Claude Opus 4.6 reproduces 30 tasks successfully ($\gimp \ge -0.05$) and DeepSeek-V4-Pro reproduces 21 tasks. On the 16 tasks where both succeed, $\gimp$ clusters tightly around zero (median $-0.0026$, $90\%$ of deviations $\le 0.031$), confirming that the SOTA anchors are well calibrated. Remaining non-successes trace to the uniform resource budget and agent capability rather than package defects (Figure~\ref{fig:nb:benchmark-calibration}).

\subsection{Benchmark Statistics}
\label{sec:nb:stats}

The benchmark comprises 90 tasks and 333 evaluation instances. We characterize NatureBench along two complementary themes: the breadth and representativeness of its coverage, and the heterogeneity of its evaluation design. The first theme describes how tasks are distributed across scientific domains, ML task types, and source-paper contribution types. The second characterizes each task's evaluation along three layers: what is evaluated (Scope), how the reference answer is defined (Paradigm), and what it is measured by (Metric). This heterogeneity explains why \S\ref{sec:nb:eval} requires a single cross-task-comparable metric.

\paragraph{Breadth and representativeness.}
Figure~\ref{fig:nb:benchmark-statistics} summarizes NatureBench coverage along three single-label axes (source journal, scientific domain, and ML task type) together with a multi-label view of the source papers' contribution nature.
By provenance, the final 90 tasks concentrate in six journals, led by \emph{Nature Machine Intelligence} (36), \emph{Nature Methods} (26), and \emph{Nature Computational Science} (16). The corpus skews recent, with 11, 17, 28, and 34 tasks for 2022 through 2025.
Across scientific domains, the tasks span six areas (cellular omics, protein biology, biomedical modeling, physical modeling, molecular design, and relational reasoning) and eight ML task types, where prediction/regression and classification dominate, followed by clustering/integration and a long tail of generation, segmentation, simulation, structure-modeling, and other specialized tasks.
Source papers also vary in contribution type: most adapt established methods to new scientific settings, a sizable share introduce algorithmic innovations, and a few contribute a new problem formulation, with a single paper often spanning more than one category.
\paragraph{Heterogeneous evaluation design.}
Figure~\ref{fig:nb:benchmark-statistics} reports the design summary.
At the \emph{Scope} layer, tasks are evaluated over multiple instances (mean 3.7, median 3, up to 19), organized under varied data-partition topologies: most use multiple independent test sets, but many use a shared training set with multiple test sets or leave-one-out cross-dataset splits, so evaluation extends beyond a single dataset to generalization conditions.
Agent-visible data ranges from under 1~GB (about half the tasks) to over 10~GB (about a fifth).
By primary input modality, the tasks span biological sequences, molecular and materials structures, single-cell and spatial omics, imaging and volumetric data, temporal signals and spectra, graphs and networks, and feature tables.
At the \emph{Paradigm} layer, most tasks use a static label scored against hidden ground truth. The remaining tasks are either \emph{distribution} tasks, where the agent generates samples scored by set-level or distributional metrics, or \emph{oracle} tasks, where the agent optimizes against a provided scorer with no fixed correct answer.
At the \emph{Metric} layer, the tasks use 81 distinct primary metrics (AUROC, RMSE, Spearman~$\rho$, ARI, F1, MAE, among others), with each task typically scored by several (mean 3.7 primary, 5.1 auxiliary), most of which are higher-is-better.
This metric heterogeneity makes per-task raw scores incomparable, motivating the direction-normalized, scale-free relative-gap metric of \S\ref{sec:nb:eval}.

\subsection{Evaluation Protocol}
\label{sec:nb:eval}

\begin{figure}[t]
\centering
\includegraphics[width=\linewidth]{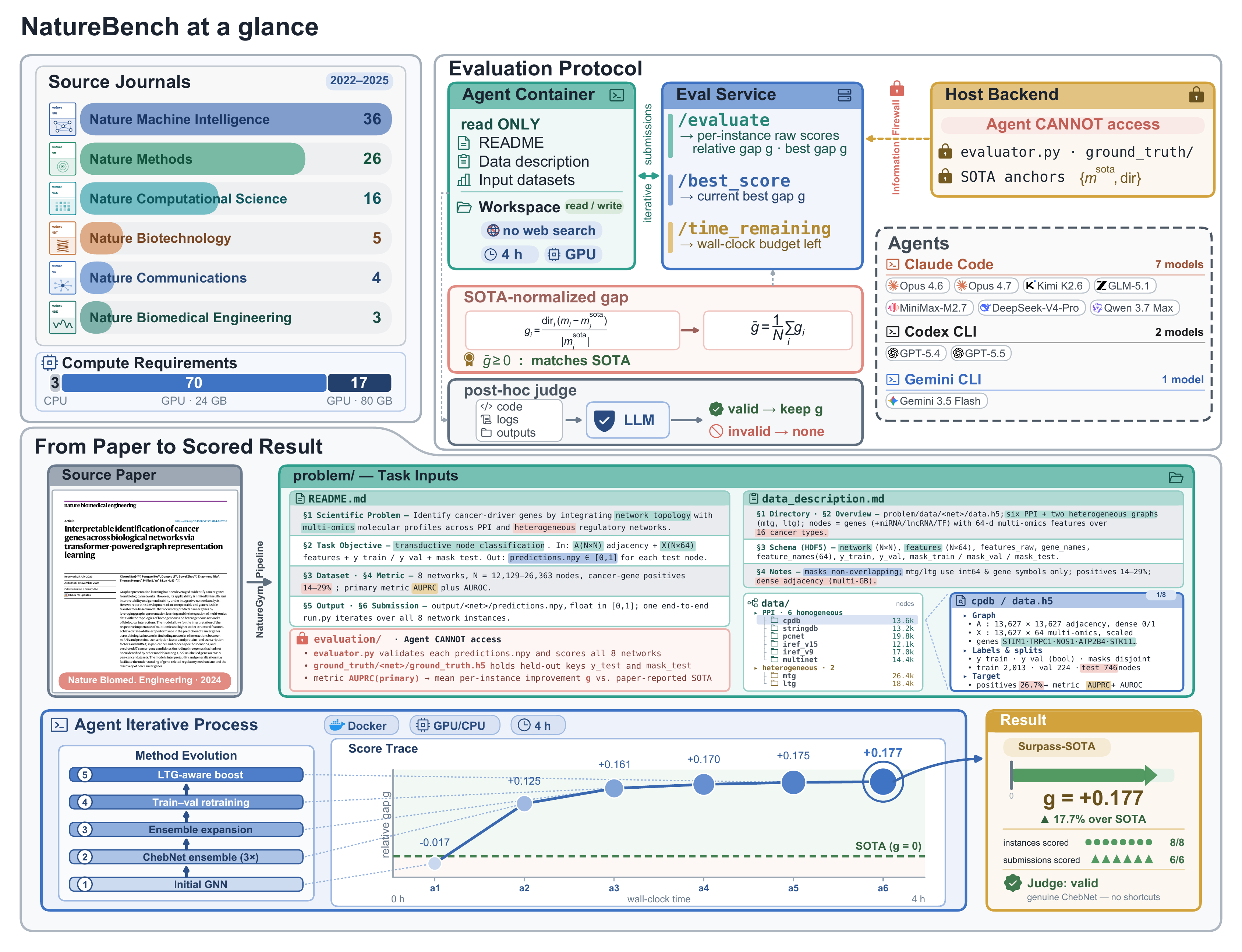}
\caption{\textbf{NatureBench task construction and evaluation pipeline.}
Each source paper is constructed into a task package that separates the agent-visible task description and data from a hidden evaluator, ground truth, and paper-reported SOTA. An agent then solves the task inside an isolated container exposing only the task description, data, and a writable workspace, while a host-side service scores each submission. A post-hoc judge screens the run for validity.}
\label{fig:nb:evaluation-protocol}
\end{figure}

Each agent solves its task inside an isolated NatureBench container, scored by a standardized evaluation service against the source paper's reported SOTA.
The protocol keeps every retained score both \emph{comparable}, because heterogeneous task metrics collapse to one SOTA-normalized quantity, and \emph{trustworthy}, because the agent is sealed from the ground truth while it works and audited for shortcuts afterwards.

\paragraph{SOTA-normalized relative gap.}
To compare agents across tasks with heterogeneous metrics, each task is scored and ranked by a single normalized quantity computed on the one primary metric that each instance designates. The remaining metrics are still reported to the agent as feedback but do not enter this normalized score.
For instance~$i$, this SOTA-normalized relative gap is
\begin{equation}
  \gimp_i \;=\; \mathrm{dir}_i \cdot \frac{m_i - m_i^{\mathrm{sota}}}{|m_i^{\mathrm{sota}}|},
\end{equation}
where $m_i$ is the agent's value on that primary metric, $m_i^{\mathrm{sota}}$ is the paper-reported SOTA for it, and $\mathrm{dir}_i \in \{+1, -1\}$ encodes the metric direction.
$\gimp_i \geq 0$ means the agent matches or surpasses the published result.
The task-level score averages $\gimp_i$ across instances, and instances with no valid submission receive $\gimp_i^{\mathrm{fail}} = -1.0$.
Because $\gimp$ is scale-free and direction-normalized, it enables direct comparison across tasks whose primary metrics are heterogeneous (e.g., AUROC, RMSE, Spearman~$\rho$).

\paragraph{Agent run and adjudication.}
The agent operates inside an isolated, task-specific Docker container with read access to \texttt{problem/} (task description and data) and read/write access to \texttt{workspace/}, a 4-hour wall-clock budget, and one GPU when the task requires it. The evaluator, ground truth, and SOTA target reside in a host-side evaluation service that the agent cannot access directly.
\emph{During} the run, the agent iteratively queries this service through three endpoints. \texttt{/evaluate} scores a submission on every instance across all reported metrics and returns raw scores, relative gaps, and the running best. \texttt{/best\_score} returns the current best without submitting. \texttt{/time\_remaining} reports the remaining budget. The wall clock pauses during scoring so that evaluation overhead does not consume the agent's budget.
\emph{After} the run, a post-hoc Claude Sonnet 4.6 judge checks for shortcut behavior (output fabrication, rule substitution for learning, answer recovery, feedback gaming, or training bypass) and assigns flagged runs a score of \texttt{none}.

\section{Experiments}
\label{sec:exp}

\subsection{Experimental Setup}
\label{sec:exp:setup}

We evaluate frontier coding agents on NatureBench under a single shared protocol, measuring how closely each approaches the published SOTA of each task's source paper. Given only a task's visible data and problem specification, an agent autonomously develops a solution and submits it iteratively, scored against the paper's SOTA target through the evaluation protocol of \S\ref{sec:nb:eval}.

\paragraph{Models.}
We evaluate twelve models, each pairing one of three CLI-based agent harnesses. Claude Code~\citep{anthropic2025claudecode} is paired with nine models: Claude Opus 4.6, Claude Opus 4.7~\citep{anthropic2026opus46,anthropic2026opus47}, Kimi K2.6~\citep{moonshot2026kimik2}, MiniMax-M2.7, MiniMax-M3~\citep{minimax2026m27,minimax2026m3}, DeepSeek-V4-Pro~\citep{deepseek2026v4}, GLM-5.1, GLM-5.2~\citep{zai2026glm51,zai2026glm52}, and Qwen 3.7 Max~\citep{qwen2026max}. Codex CLI~\citep{openai2025codex} is paired with GPT-5.4 and GPT-5.5~\citep{openai2026gpt54,openai2026gpt55}. Gemini CLI~\citep{google2025geminicli} is paired with Gemini 3.5 Flash~\citep{google2026gemini35flash}. Each agent is run independently over all 90 tasks.

\paragraph{Unified conditions.}
All agents disable web search, preventing them from retrieving the source dataset or paper content as a shortcut. Each harness keeps its default reasoning-effort setting. Every task is given the same 4-hour wall-clock budget and a GPU matched to the compute requirement recorded in its metadata (\S\ref{sec:gym:build}): the 3 tasks needing no GPU run CPU-only, the 70 with lighter GPU requirements each receive a single NVIDIA RTX 3090 or 4090, and the 17 most compute-intensive receive a single NVIDIA A800. All the evaluation mechanics follow the protocol of \S\ref{sec:nb:eval}. Appendix~\ref{app:cost} reports per-agent token and turn statistics.

\begin{table}[t]
\centering
\caption{\textbf{Main results on NatureBench,} sorted by overall Surpass-SOTA. Each group reports \textbf{S}\,=\,Surpass-SOTA ($g > 0.1$) and \textbf{M}\,=\,Match-SOTA ($g \ge 0$), as percentages of tasks, both overall (All) and per scientific domain. Best/second in the All columns are \textbf{bold}/\underline{underlined}.}
\label{tab:main-results}
\resizebox{\textwidth}{!}{%
\begin{tabular}{l cc cc cc cc cc cc cc}
\toprule
 & \multicolumn{2}{c}{All} & \multicolumn{2}{c}{Protein} & \multicolumn{2}{c}{Cellular} & \multicolumn{2}{c}{Physical} & \multicolumn{2}{c}{Molec.} & \multicolumn{2}{c}{Relat.} & \multicolumn{2}{c}{Biomed.} \\
\cmidrule(lr){2-3}\cmidrule(lr){4-5}\cmidrule(lr){6-7}\cmidrule(lr){8-9}\cmidrule(lr){10-11}\cmidrule(lr){12-13}\cmidrule(lr){14-15}
Model & S\,$\uparrow$ & M\,$\uparrow$ & S\,$\uparrow$ & M\,$\uparrow$ & S\,$\uparrow$ & M\,$\uparrow$ & S\,$\uparrow$ & M\,$\uparrow$ & S\,$\uparrow$ & M\,$\uparrow$ & S\,$\uparrow$ & M\,$\uparrow$ & S\,$\uparrow$ & M\,$\uparrow$ \\
\midrule
Claude Opus 4.7   & \hsb{17.8} & \hmb{47.8} & \ds{12.5} & \dm{56.2} & \ds{22.6} & \dm{54.8} & \ds{30.8} & \dm{46.2} & \ds{18.2} & \dm{45.5} & \ds{0.0} & \dm{60.0} & \ds{7.1} & \dm{21.4} \\
GLM-5.2           & \hsu{15.6} & \hm{41.1} & \ds{12.5} & \dm{43.8} & \ds{25.8} & \dm{51.6} & \ds{23.1} & \dm{23.1} & \ds{0.0} & \dm{45.5} & \ds{0.0} & \dm{60.0} & \ds{7.1} & \dm{21.4} \\
Gemini 3.5 Flash  & \hsu{15.6} & \hm{37.8} & \ds{6.2} & \dm{43.8} & \ds{25.8} & \dm{51.6} & \ds{30.8} & \dm{30.8} & \ds{0.0} & \dm{18.2} & \ds{0.0} & \dm{60.0} & \ds{7.1} & \dm{14.3} \\
GPT-5.5           & \hs{14.4} & \hmu{44.4} & \ds{6.2} & \dm{50.0} & \ds{25.8} & \dm{54.8} & \ds{23.1} & \dm{38.5} & \ds{0.0} & \dm{18.2} & \ds{0.0} & \dm{60.0} & \ds{7.1} & \dm{35.7} \\
Claude Opus 4.6   & \hs{12.2} & \hm{36.7} & \ds{12.5} & \dm{31.2} & \ds{19.4} & \dm{41.9} & \ds{23.1} & \dm{30.8} & \ds{0.0} & \dm{36.4} & \ds{0.0} & \dm{60.0} & \ds{0.0} & \dm{28.6} \\
MiniMax-M3        & \hs{11.1} & \hm{33.3} & \ds{12.5} & \dm{43.8} & \ds{19.4} & \dm{35.5} & \ds{15.4} & \dm{30.8} & \ds{0.0} & \dm{36.4} & \ds{0.0} & \dm{60.0} & \ds{0.0} & \dm{7.1} \\
Qwen 3.7 Max      & \hs{10.0} & \hm{28.9} & \ds{12.5} & \dm{37.5} & \ds{16.1} & \dm{35.5} & \ds{15.4} & \dm{23.1} & \ds{0.0} & \dm{18.2} & \ds{0.0} & \dm{40.0} & \ds{0.0} & \dm{14.3} \\
Kimi K2.6         & \hs{8.9} & \hm{30.0} & \ds{12.5} & \dm{37.5} & \ds{12.9} & \dm{29.0} & \ds{15.4} & \dm{15.4} & \ds{0.0} & \dm{27.3} & \ds{0.0} & \dm{60.0} & \ds{0.0} & \dm{28.6} \\
GPT-5.4           & \hs{8.9} & \hm{27.8} & \ds{6.2} & \dm{37.5} & \ds{12.9} & \dm{29.0} & \ds{23.1} & \dm{30.8} & \ds{0.0} & \dm{18.2} & \ds{0.0} & \dm{60.0} & \ds{0.0} & \dm{7.1} \\
GLM-5.1           & \hs{7.8} & \hm{28.9} & \ds{6.2} & \dm{25.0} & \ds{12.9} & \dm{35.5} & \ds{7.7} & \dm{23.1} & \ds{0.0} & \dm{18.2} & \ds{0.0} & \dm{60.0} & \ds{7.1} & \dm{21.4} \\
DeepSeek-V4-Pro   & \hs{4.4} & \hm{26.7} & \ds{6.2} & \dm{37.5} & \ds{9.7} & \dm{32.3} & \ds{0.0} & \dm{15.4} & \ds{0.0} & \dm{18.2} & \ds{0.0} & \dm{60.0} & \ds{0.0} & \dm{7.1} \\
MiniMax-M2.7      & \hs{1.1} & \hm{13.3} & \ds{0.0} & \dm{18.8} & \ds{3.2} & \dm{16.1} & \ds{0.0} & \dm{7.7} & \ds{0.0} & \dm{0.0} & \ds{0.0} & \dm{20.0} & \ds{0.0} & \dm{14.3} \\
\bottomrule
\end{tabular}%
}
\end{table}

\subsection{Main Results}
\label{sec:exp:main}

Clear improvements over the published SOTA are rare across all twelve agents, and even the best matches it on fewer than half of the 90 tasks. Table~\ref{tab:main-results} reports Surpass-SOTA ($g > 0.1$) and Match-SOTA ($g \ge 0$) rates, both overall and per scientific domain.

\paragraph{Overall performance.}
Clear improvements over the published SOTA ($g > 0.1$) are uncommon even for the strongest agents: Claude Opus 4.7 reaches only $17.8\%$, followed by Gemini 3.5 Flash and GLM-5.2 (both $15.6\%$), and GPT-5.5 ($14.4\%$), while MiniMax-M2.7 falls to $1.1\%$. Match-SOTA rates ($g \ge 0$) are higher but still below half: Claude Opus 4.7 leads at $47.8\%$, followed by GPT-5.5 ($44.4\%$), GLM-5.2 ($41.1\%$), and Gemini 3.5 Flash ($37.8\%$). The remaining agents range from $13.3\%$ to $36.7\%$, with MiniMax-M2.7 trailing. The per-domain columns of Table~\ref{tab:main-results} show that attainment is distributed unevenly across scientific domains, and that clear improvements are more concentrated. We defer this cross-domain structure to \S\ref{sec:exp:domain}.

\begin{table}[t]
\centering
\caption{\textbf{Gap summary and submission rates of agents on NatureBench.} The SOTA-normalized gap $g$ (\S\ref{sec:nb:eval}) is summarized by its median $\tilde{g}$ and mean $\bar{g}$; the $\cdot_{\text{all}}$ columns set $g=-1.0$ for tasks with no valid score, while the $\cdot_{\text{valid}}$ columns cover only judge-accepted tasks. \textbf{CR} (Completion Rate) and \textbf{SR} (Score Rate) are the fractions of tasks yielding a valid score and any score. Best/second per column in \textbf{bold}/\underline{underlined}.}
\label{tab:gap-reliability}
\resizebox{\textwidth}{!}{%
\begin{tabular}{l l cccc cc}
\toprule
 & & \multicolumn{4}{c}{Gap Summary} & \multicolumn{2}{c}{Submission Rates (\%)} \\
\cmidrule(lr){3-6}\cmidrule(lr){7-8}
Model & Harness & $\tilde{g}_{\text{all}}$ & $\bar{g}_{\text{all}}$ & $\tilde{g}_{\text{valid}}$ & $\bar{g}_{\text{valid}}$ & CR & SR \\
\midrule
Claude Opus 4.7   & Claude Code & \mgb{-0.007} & \mn{-4.54}  & \mgu{-0.007} & \mn{-4.54}  & \relb{100.0} & \relb{100.0} \\
GLM-5.2           & Claude Code & \mg{-0.058}  & \mn{-3.85}  & \mg{-0.051}  & \mn{-3.95}  & \rel{96.7}   & \relu{98.9}  \\
Gemini 3.5 Flash  & Gemini CLI  & \mg{-0.083}  & \mn{-5.71}  & \mg{-0.041}  & \mn{-5.98}  & \rel{94.4}   & \relu{98.9}  \\
GPT-5.5           & Codex CLI   & \mgu{-0.055} & \mnu{-2.81} & \mgb{+0.001} & \mn{-3.14}  & \rel{84.4}   & \relu{98.9}  \\
Claude Opus 4.6   & Claude Code & \mg{-0.061}  & \mnb{-2.02} & \mg{-0.061}  & \mnb{-2.02} & \relb{100.0} & \relb{100.0} \\
MiniMax-M3        & Claude Code & \mg{-0.103}  & \mn{-5.90}  & \mg{-0.100}  & \mn{-5.95}  & \relu{98.9}  & \relu{98.9}  \\
Qwen 3.7 Max      & Claude Code & \mg{-0.121}  & \mn{-2.94}  & \mg{-0.105}  & \mnu{-3.03} & \rel{95.6}   & \relu{98.9}  \\
Kimi K2.6         & Claude Code & \mg{-0.142}  & \mn{-10.11} & \mg{-0.087}  & \mn{-10.88} & \rel{92.2}   & \rel{94.4}   \\
GPT-5.4           & Codex CLI   & \mg{-0.123}  & \mn{-3.72}  & \mg{-0.113}  & \mn{-3.88}  & \rel{94.4}   & \relb{100.0} \\
GLM-5.1           & Claude Code & \mg{-0.150}  & \mn{-8.44}  & \mg{-0.131}  & \mn{-8.98}  & \rel{93.3}   & \rel{93.3}   \\
DeepSeek-V4-Pro   & Claude Code & \mg{-0.242}  & \mn{-8.57}  & \mg{-0.239}  & \mn{-8.66}  & \relu{98.9}  & \relu{98.9}  \\
MiniMax-M2.7      & Claude Code & \mg{-0.401}  & \mn{-11.76} & \mg{-0.347}  & \mn{-12.53} & \rel{93.3}   & \relu{98.9}  \\
\bottomrule
\end{tabular}%
}
\end{table}

\begin{figure}[t]
\centering
\includegraphics[width=\linewidth]{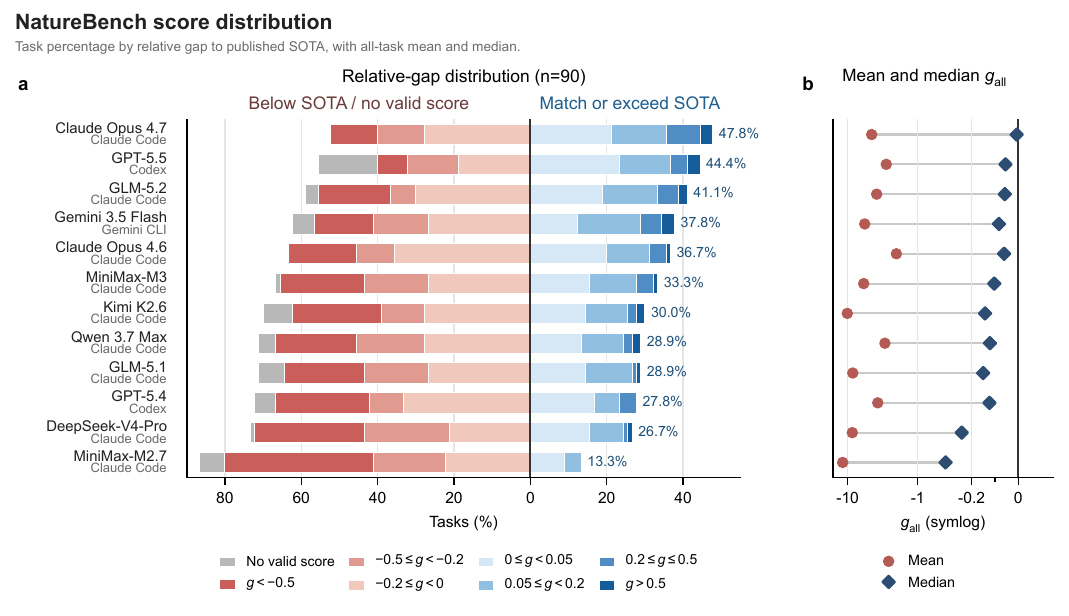}
\caption{\textbf{Gap distribution and summary of agents on NatureBench.} \textbf{a}, Percentage of tasks in each $g$ interval for each agent, arranged around the SOTA target ($g = 0$). \textbf{b}, Mean and median relative gap over all tasks ($g_{\text{all}}$), assigning $g=-1.0$ when no valid score is available.}
\label{fig:score-distribution}
\end{figure}

\paragraph{Completion and validity.}
Agents submit a scorable solution on nearly all tasks, and the few invalid shortcut submissions are filtered by the validity judge. In Table~\ref{tab:gap-reliability}, the gap between SR and CR isolates scored-but-invalid (shortcut) submissions flagged by the validity judge. The two Claude Opus agents are the cleanest, with $100\%$ on both rates and no invalid submissions, so their unmatched tasks reflect genuine performance shortfalls rather than invalid methods. GPT-5.5 attempts shortcuts most often, with $13$ invalid submissions. Because these are filtered from its score, its second-highest Match-SOTA ($44.4\%$) and the only non-negative median over judge-accepted tasks ($\tilde{g}_{\text{valid}} = +0.001$) remain genuine. Among the remaining agents, GLM-5.1 has the lowest SR ($93.3\%$): on the tasks it leaves unscored, the agent's own solution never produces a scorable submission.

\paragraph{Score distribution.}
Most tasks land modestly below SOTA rather than reaching it or failing badly. The median relative gap $\tilde{g}_{\text{all}}$ ranges from $-0.007$ for the strongest agent, Claude Opus 4.7, to $-0.40$ for the weakest, MiniMax-M2.7 (Table~\ref{tab:gap-reliability}). Figure~\ref{fig:score-distribution} shows the full spread: each agent's scores center in this moderate sub-SOTA range, with the weaker agents shifting more mass into severe failure and only a minority of tasks on any agent reaching SOTA. A few tasks carry extreme negative values because the SOTA-normalized gap amplifies large shortfalls, pulling every agent's mean far below its median. We therefore treat Surpass-SOTA and Match-SOTA as the primary metrics and the median as an auxiliary summary. \S\ref{sec:exp:validity} confirms that these extreme values reflect normalization effects rather than faulty tasks.

\section{Analysis}
\label{sec:analysis}

Agents remain far from paper-reported SOTA; we now ask how that gap arises, where it concentrates, and how reliably it is measured. This detailed behavioral analysis covers ten agent configurations. The gap is primarily one of method: agents succeed mainly by recasting scientific tasks as generic ML pipelines rather than by genuine scientific discovery, and fail mostly at method choice and execution depth (\S\ref{sec:exp:discovery-modes}). It concentrates by task: the six scientific domains form a difficulty gradient shared across the ten analyzed agents, and cross-discipline tasks widen the gap further (\S\ref{sec:exp:domain}). And it is measured reliably: extreme scores are legitimate outputs of the SOTA-relative gap, leakage- or gaming-prone tasks are caught by the protocol, and the narrowed coverage of each source paper is immaterial: we evaluate each paper's core task rather than reproduce it in full (\S\ref{sec:exp:validity}).

\subsection{Solution Mechanisms}
\label{sec:exp:discovery-modes}

\begin{figure}[t]
\centering
\includegraphics[width=\linewidth]{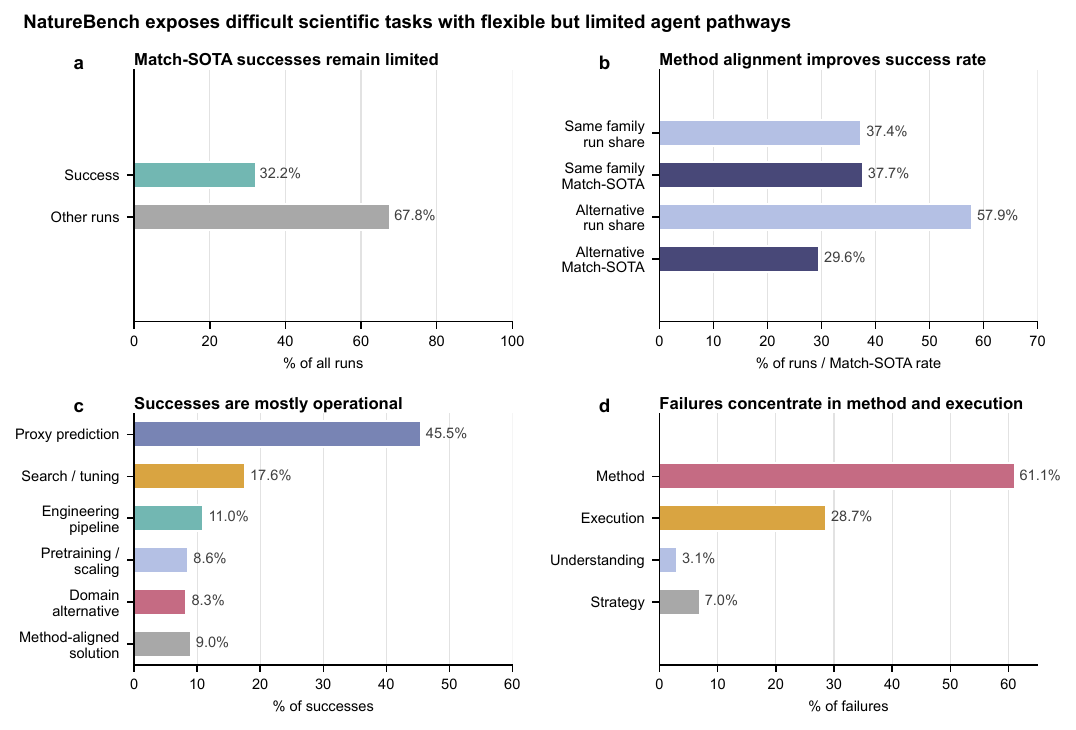}
\caption{\textbf{Solution mechanisms of ten analyzed agents across $900$ NatureBench runs.}
\textbf{a}, Match-SOTA outcomes across all runs.
\textbf{b}, Match-SOTA rates for runs using the same versus different broad method families as the source paper.
\textbf{c}, Success-mode distribution among Match-SOTA runs.
\textbf{d}, Failure-layer distribution among below-SOTA and invalid runs.}
\label{fig:solution-mechanisms}
\end{figure}

To understand not just whether agents match SOTA but how they succeed or fail, we annotate $900$ runs ($90$ tasks $\times$ $10$ analyzed agents) by comparing the paper-side method family with the agent's implemented method, attributing Match-SOTA runs to success modes, and categorizing below-SOTA or invalid runs into failure layers. As shown in Fig.~\ref{fig:solution-mechanisms}a, the Match-SOTA rate across these analyzed agents is only $32.2\%$, and this is primarily because matching published SOTA requires both choosing methods that fit the scientific structure of the task and executing them deeply enough.

\paragraph{Method pathways.}
Agents systematically reshape scientific tasks into more familiar method families: while paper-side methods concentrate in structured representation, statistical modeling, and pretraining or transfer learning, agent-side methods are concentrated in supervised predictive modeling ($41.4\%$ of runs). These shifts are not equally effective, however. As shown in Fig.~\ref{fig:solution-mechanisms}b, runs whose agent method falls into the same broad family as the source paper match SOTA in $37.7\%$ of cases, compared with $29.6\%$ for runs using a different family. Although NatureBench imposes no constraint on method choice, methods closer to the task's original scientific structure tend to be more effective.

\paragraph{Success modes.}
When agents do match SOTA, they usually do so through generic ML engineering rather than domain-informed methodological choices. As shown in Fig.~\ref{fig:solution-mechanisms}c, supervised proxy prediction accounts for $45.5\%$ of successful runs, optimization and tuning for $17.6\%$, engineering pipelines for $11.0\%$, and pretraining or model scaling for $8.6\%$. Together, these engineering-driven categories account for $82.7\%$ of successes. In contrast, domain-reasoned alternatives and method-aligned solutions account for only $8.3\%$ and $9.0\%$, respectively. This pattern suggests that agents predominantly succeed by reducing scientific tasks to standard ML pipelines (trainable, tunable, and engineerable) rather than by reasoning about the task's scientific specifics.

\paragraph{Failure modes.}
Most failures stem from method choice or execution depth, not from misunderstanding the task or producing malformed output. Among the $67.8\%$ of runs that fall below Match-SOTA or lack a valid score, method-layer failures dominate at $61.1\%$, primarily wrong method choice ($45.1\%$), followed by execution-layer failures at $28.7\%$, largely due to insufficient budget or time ($24.4\%$). Understanding-layer and strategy-layer failures account for only $3.1\%$ and $7.0\%$, respectively (Fig.~\ref{fig:solution-mechanisms}d). Most of these runs do produce runnable solutions, but the chosen method is too weak or the implementation too shallow to close the gap to paper-reported SOTA. The failure distribution thus indicates that method selection and implementation depth, rather than code generation itself, are the primary bottlenecks for current agents on NatureBench tasks. Appendix~\ref{app:trajectory-cases} presents three representative trajectories illustrating these patterns.

\subsection{Domain and Interdisciplinary Variation}
\label{sec:exp:domain}

\begin{figure}[t]
\centering
\includegraphics[width=\linewidth]{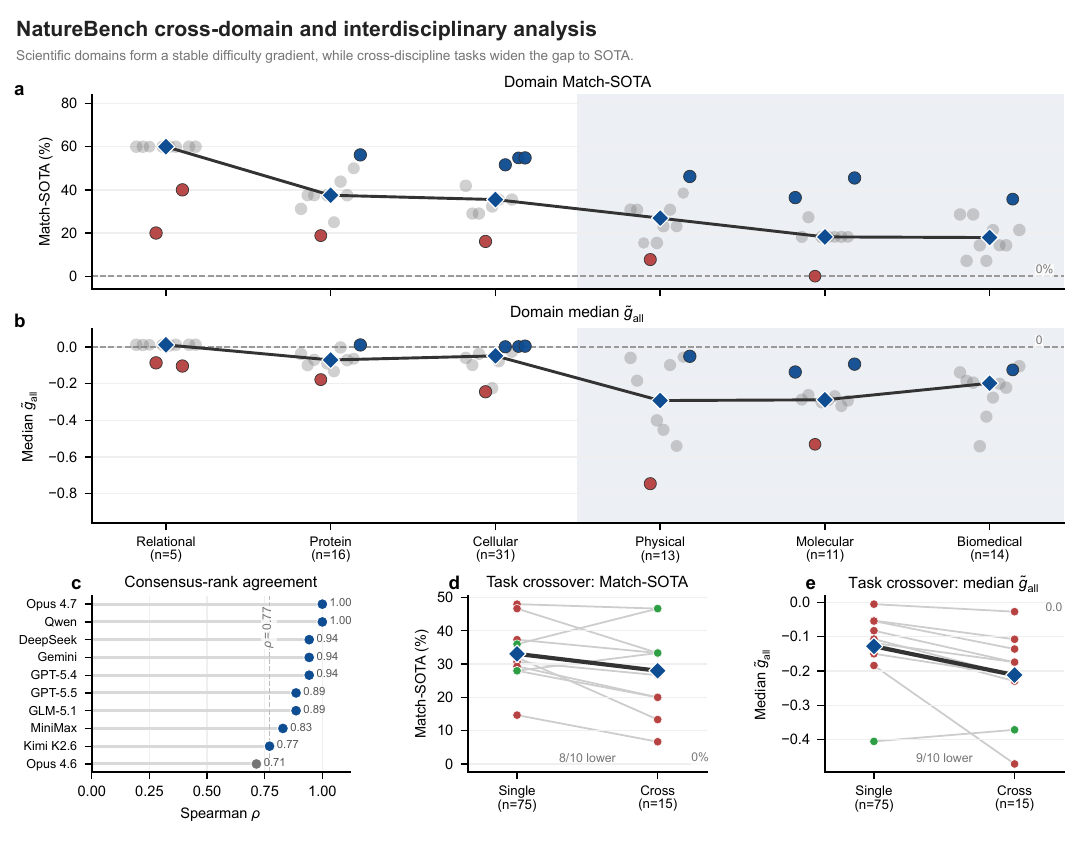}
\caption{\textbf{NatureBench performance by scientific domain and disciplinary scope.}
\textbf{a,b,} Match-SOTA rate and median $\tilde{g}_{\mathrm{all}}$ across six domains for $10$ analyzed agents. Grey circles: agents; blue diamonds: domain medians. Blue/red circles mark higher/lower deviations from those medians (at least $15$ percentage points in the Match-SOTA rate and a same-direction $\tilde{g}_{\mathrm{all}}$ shift).
\textbf{c,} Spearman $\rho$ between each agent's domain ranking and the consensus Match-SOTA ranking.
\textbf{d,e,} The same metrics on $75$ single- versus $15$ cross-discipline tasks. Diamonds: across-agent means; green/red pairs: increases/decreases.}
\label{fig:cross-domain-performance}
\end{figure}

We examine whether the scientific domain and disciplinary scope of a task systematically affect agent performance. Both factors prove influential: the six domains form a stable difficulty gradient, with the consensus Match-SOTA rate ranging from $60.0\%$ down to $17.9\%$, and this ordering is highly consistent across all ten agents ($\rho \ge 0.71$). Interdisciplinary tasks further widen the gap to paper-reported SOTA. Figure~\ref{fig:cross-domain-performance} presents the full decomposition.

\paragraph{Scientific domain.} Performance varies across the six scientific domains, and this difficulty ordering is shared across agents.
Ranking the six domains by the consensus Match-SOTA rate reveals a difficulty gradient that separates into two tiers. The easier tier comprises Relational Reasoning ($60.0\%$), Protein Biology ($37.5\%$), and Cellular Omics ($35.5\%$). The harder tier comprises Physical Modeling ($26.9\%$), Molecular Design ($18.2\%$), and Biomedical Modeling ($17.9\%$). The consensus $\tilde{g}_{\text{all}}$ corroborates this split: the median relative gap stays within $8\%$ for the easier tier ($\tilde{g} > -0.08$) but exceeds $20\%$ for the harder tier ($\tilde{g} < -0.20$). All ten agents rank-correlate positively with this ordering (Spearman $\rho$ from $0.71$ to $1.00$, nine at $\rho \ge 0.77$), indicating that this cross-domain variation is largely shared across agents rather than specific to any individual agent.

\paragraph{Interdisciplinary tasks.}
Beyond performance spread across the six domains, a subset of tasks each integrate more than one domain within a single task, and these tend to be solved further from SOTA than single-discipline tasks. We tag each task by whether it draws on more than one scientific domain, yielding $15$ cross-discipline and $75$ single-discipline tasks. Comparing the two groups, we find that the pooled median $\tilde{g}_{\text{all}}$ falls from $-0.13$ on single-discipline tasks to $-0.21$ on cross-discipline tasks, with $9$ of $10$ agents moving in this direction. The Match-SOTA rate shows the same direction, dropping from $33.1\%$ to $28.0\%$, with $8$ of $10$ agents lower. The consistent widening of the agent--SOTA gap on interdisciplinary tasks suggests that integrating knowledge across domains remains a distinct challenge for most current agents.

\subsection{Benchmark Validity}
\label{sec:exp:validity}

NatureBench converts public papers into automatically scored tasks and normalizes their heterogeneous metrics onto a common SOTA-relative scale. To verify that this design does not distort the results, we audit the tasks with extreme scores and those most exposed to leakage or gaming. We examined each concern and found it either working as designed or bounded to acceptable levels by the protocol.

\paragraph{Metric normalization.}
Extreme scores are a property of the SOTA-relative metric rather than a sign of a faulty task, surfacing as the heavy negative tail in Fig.~\ref{fig:score-distribution} and the gap between $\bar{g}$ and $\tilde{g}$ in Table~\ref{tab:gap-reliability}. The gap $g=(\text{score}-\text{SOTA})/|\text{SOTA}|$ scores each result as a fraction of the reported SOTA, so its magnitude depends on that SOTA as much as on the agent. A near-ceiling SOTA leaves a tiny denominator, so a merely moderate agent maps to a large negative $g$ on a genuinely hard task. A large positive $g$ may arise where the single primary metric used for scoring captures only one facet of a multi-objective method that its source paper evaluates with several metrics across different aspects: an agent optimizing for it directly can exceed the reported value without pursuing the method's other objectives. Auditing every extreme-gap task, we find no task error. We therefore use Surpass- and Match-SOTA as the primary metrics and the median $\tilde{g}$ as a tail-robust summary, with the mean only for completeness.

\paragraph{Task coverage.}
Some tasks evaluate only a bounded slice of their source paper, an unavoidable and reasonable narrowing. Each such task retains the paper's core quantitative problem and scores a subset of instances and metrics. When the omitted instances or metrics cover other directions of the contribution, the paper is captured only in part. A direction is usually excluded because it cannot be captured as structured data or scored automatically and deterministically. Separately, obtainable instances past a task's data-volume budget are also not collected. The retained slice is still the paper's core quantitative task and is scored correctly, so Surpass- and Match-SOTA measure performance on that slice, not on the whole paper.

\paragraph{Leakage and feedback.}
The residual leakage and feedback risks are unavoidable but constrained by the protocol and confirmed bounded by review. Because tasks are built from public data, some information is in principle accessible: source datasets come from public repositories and benchmarks, and on a few tasks the agent-visible inputs are inherently coupled to their targets, so an agent might read off part of the answer rather than compute it. A secondary risk is that exact-score feedback over repeated submissions lets an agent game the scorer rather than solve the task. The protocol bounds both: web search is disabled, so agents cannot retrieve the data or reported results, and a post-hoc validity judge filters scored-but-invalid submissions (the SR--CR gap in Table~\ref{tab:gap-reliability}). Reviewing the most at-risk tasks, we find high-frequency submission is overwhelmingly legitimate iteration, and the rare genuine exploit is caught by the judge.

\section{Related Work}
\label{sec:related}

\subsection{AI for Science}
\label{sec:rel:ai4sci}
\paragraph{The first wave: AI as an accelerator within human-defined research programs.}
AI for Science has produced strong vertical results across many disciplines. In structural biology, AlphaFold, RoseTTAFold, ESMFold, AlphaFold~3, and Boltz-1 expand atomic-level prediction from single chains to biomolecular complexes~\citep{jumper2021highly,baek2021accurate,lin2023evolutionary,abramson2024accurate,wohlwend2025boltz}, while RFdiffusion and its antibody extension close the loop with experimentally validated \emph{de novo} design~\citep{watson2023novo,bennett2026atomically}. In genomics, AlphaMissense and PheMART model variant pathogenicity and phenotype space~\citep{cheng2023accurate,wen2026phenotypic}. Geneformer, scGPT, and Evo~2 pretrain foundation models over transcriptomes or DNA~\citep{theodoris2023transfer,cui2024scgpt,brixi2026genome}. Cell2location resolves cell types in spatial transcriptomics~\citep{kleshchevnikov2022cell2location}. In materials, chemistry, mathematics, and Earth systems, GNoME and MatterGen discover or inverse-design materials~\citep{merchant2023scaling,zeni2025generative}, Coscientist automates chemical experimentation~\citep{boiko2023autonomous}, AlphaTensor and AlphaProof extend search-based reasoning to algorithms and formal mathematics~\citep{fawzi2022discovering,hubert2025olympiad}, and GraphCast, GenCast, and Aurora advance global weather and Earth-system prediction~\citep{lam2023learning,price2025probabilistic,bodnar2025foundation}.

\paragraph{A structural limitation of the \emph{research-plus-AI} paradigm.}
Powerful as these systems are, they mostly share the same methodological form: humans specify the research programme, curate the data, and fix the success criterion, while AI acts as a more capable instrument inside that programme. This makes many advances a \emph{revolution of tools} rather than a \emph{tool of revolution}~\citep{zhou2025advancing}. Large-scale publication evidence further suggests that AI-augmented science can raise individual output and impact while narrowing the collective topic frontier toward data-rich subfields~\citep{hao2026artificial}. Thus, existing AI-for-Science systems can accelerate progress along established axes, but they do not by themselves establish cross-disciplinary, paradigm-shifting problem solving.

\paragraph{From AI-assisted research to AI-native problem solving.}
The natural next step is to evaluate AI as the primary problem solver: given a scientific task, the system must choose methods, run experiments, and be judged by the final scientific outcome. General-purpose scientific agents such as The AI Scientist, the AI co-scientist, DeepScientist, and AutoSOTA move in this direction~\citep{lu2026towards,gottweis2026accelerating,weng2025deepscientist,li2026autosota}, but they are usually demonstrated on self-selected topics or within limited domains, leaving open whether AI-native problem solving generalizes across science as a whole.

\paragraph{Cross-disciplinary evaluation as a test of breaking the information cocoon.}
Contemporary scientists face an increasingly restrictive information cocoon: specialized training, literature growth, and field-specific tooling make it difficult to integrate methods, data, and concepts across disciplines~\citep{hao2026artificial,zhou2025advancing,piao2023human}. This is where an AI-native solver should have a distinctive advantage, because the same agent can combine biological representation learning, chemical search, physical simulation, and statistical modeling within one system. NatureBench therefore tests the missing horizontal capability: whether contemporary coding agents can solve 90 Nature-family tasks across six scientific task domains, using each paper's reported SOTA as a unified discovery scoring anchor and evaluating whether agents can move beyond field-specific \emph{research-plus-AI} toward cross-disciplinary scientific problem solving.

\subsection{Paper-based Benchmarks}
\label{sec:rel:paper}
The paper-based benchmark literature asks whether agents can read, evaluate, and operationalize scientific papers as the core artifact. One line targets paper understanding: PaperQA, PaperQA2, and OpenScholar evaluate retrieval-augmented, citation-backed answers or literature syntheses~\citep{lala2023paperqa,skarlinski2024paperqa2,asai2024openscholar}. LAB-Bench extends this to biology papers with supplementary materials, figures, tables, and protocols~\citep{laurent2024labbench}. ReviewerGPT, large-scale LLM-feedback studies, and MMReview test peer-review-style critique over text-only, multidisciplinary, or multimodal manuscripts~\citep{liu2023reviewergpt,liang2023llmfeedback,gao2025mmreview}.

A second line turns papers into executable work: PaperBench asks agents to reconstruct ICML papers from scratch under author-informed rubrics~\citep{starace2025paperbench}, while AutoExperiment and LMR-Bench use progressive code masking or language-modeling research specifications to test recovery of reported experiments~\citep{kim2025autoexperiment,yan2025lmrbench}. Reproducibility benchmarks broaden this beyond ML: CORE-Bench, REPRO-Bench, and ReplicationBench cover reproduction, assessment, or replication across computer science, social science, medicine, and astrophysics~\citep{siegel2024corebench,hu2025reprobench,ye2025replicationbench}. AutoMat and Collider-Bench add materials-science and Large Hadron Collider toolchains~\citep{huang2026automat,faroughy2026colliderbench}. FIRE-Bench asks agents to rediscover established insights from high-level questions extracted from ML papers~\citep{wang2026firebench}. These benchmarks ground evaluation in papers, but their target is reading, review, reproduction, replication, reproducibility assessment, or rediscovery of known findings. NatureBench keeps paper grounding while shifting the target to independently solving the same scientific problem, using the source paper's reported SOTA as the scoring anchor to match or surpass.

\subsection{AI-train-AI and Autonomous Optimization}
\label{sec:rel:mle}

Recent AI-train-AI, autonomous-optimization, and auto-research work can be organized by how it models the agent's task. Benchmark suites such as MLAgentBench, MLE-bench, MLGym, MLE-Dojo, MLS-Bench, AIRS-Bench, PostTrainBench, InferenceBench, and AutoLab evaluate agents over collections of ML experimentation, model-building, post-training, inference-optimization, or long-horizon closed-loop optimization tasks~\citep{huang2023mlagentbench,shern2024mlebench,nathani2025mlgym,qiang2026mle,lyu2026mlsbench,lupidi2026airsbench,rank2026posttrainbench,yeon2026inferencebench,xu2026autolab}. FrontierCS, ALE-Bench, and Frontier-Eng extend this suite-style evaluation to algorithm engineering and real-world engineering optimization~\citep{mang2025frontiercs,imajuku2025alebench,chi2026frontiereng}. A second line studies few-task, verifier-driven discovery, where agents repeatedly propose, execute, and evaluate programs, algorithms, or scientific candidates on specialized high-value objectives~\citep{fawzi2022alphatensor,romeraparedes2023funsearch,novikov2025alphaevolve,wang2025thetaevolve,yuksekgonul2026tttdiscover,ye2026simpletes,cemri2026adaevolve,liu2026evox,jiang2026deltaevolve,lin2026aevolve,liu2025deepevolve}. A third line frames the task as end-to-end research automation, including simulated scientific environments, autonomous paper-generation workflows, multi-agent hypothesis generation, lab-in-the-loop discovery, SOTA model discovery, and reviewer-style evaluation of generated research~\citep{jansen2024discoveryworld,lu2024aiscientist,gottweis2026coscientist,ghareeb2026robin,li2026autosota,weng2025deepscientist,zhang2026researcharena,lyu2026evoscientist,zhu2026evomaster}. These task models leave the key intersection underexplored: large-scale benchmark suites grounded in paper-level scientific research and evaluated against the paper's reported SOTA on its core scientific metric. NatureBench fills this gap with 90 Nature-family tasks that combine benchmark-suite scale, paper-sourced science, and SOTA-referenced evaluation across six scientific task domains.

\section{Conclusion}
\label{sec:conclusion}
 
\noindent
We introduced NatureGym, an automated pipeline that constructs per-task scientific environments from Nature-family papers, and NatureBench, a benchmark of $90$ Nature-sourced tasks across six scientific domains that uses these environments to measure not just reproduction but \emph{discovery}.
Across ten frontier agents, the strongest surpasses the published SOTA ($g > 0.1$) on only $17.8\%$ of tasks and matches it on $47.8\%$.
The dominant success pathway is methodological translation, where agents convert scientific tasks into familiar supervised-prediction problems, rather than scientific invention. Failures are dominated by wrong method choice ($45.1\%$) and insufficient compute budget ($24.4\%$), not by task misunderstanding.
We release \benchmark, NatureGym, and a public leaderboard with maintainer-side reproduction, with the long-term aim of turning the same substrate into training data for future scientific-discovery agents.

\section{Authors}
\label{app:contributions}

\textbf{Core Authors}

Yuru Wang$^{1, 2}$, Lejun Cheng$^3$, Yuxin Zuo$^2$

\textbf{Contributors}

Sihang Zeng$^{4}$, Bingxiang He$^{2}$, Che Jiang$^{1, 2}$, Junlin Yang$^{1, 2}$, Yuchong Wang$^{1, 2}$, Kaikai Zhao$^{2}$ \\
Weifeng Huang$^{2}$, Kai Tian$^{1, 2}$, Zhenzhao Yuan$^{1, 2}$, Jincheng Zhong$^{1, 2}$, Weizhi Wang$^{1, 2}$

\textbf{Corresponding Authors}

Ning Ding$^{2}$, Bowen Zhou$^{2}$, Kaiyan Zhang$^{1}$

\textbf{Main Affiliations}

$^1$ Horizon Research, Frontis.AI \quad $^2$ Tsinghua University \\
$^3$ Peking University \quad $^4$ Harvard University

\clearpage
\bibliography{references}

\appendix

\section{Package and Environment Review Details}
\label{app:package-review}

This appendix expands the package and environment review summarized in \S\ref{sec:gym:build}. Unlike the one-shot reviews of the filtering and data-acquisition stages, this review runs a verify--repair loop that iterates until the final artifact is structurally complete, internally consistent, stably scorable by the evaluator, and buildable into a working environment, all while preserving the information firewall. It has three parts.

\paragraph{Build-time self-audit.}
Before completing the construction, a final step re-reads the paper and the structured record to recheck the task definition, data alignment, metadata tags, SOTA scores, and the firewall. Anything the automated process is uncertain about is flagged for human review before the loop proceeds.

\paragraph{Task-package verification.}
We run 36 checks across five dimensions: artifact completeness, cross-component consistency, the information firewall, benchmark-design principles, and end-to-end dynamic testing. The dynamic test runs a simple baseline solver that follows the \texttt{README} interface end to end over all instances and feeds its outputs to the evaluator, checking that the score structure and values are sensible, and adds a correctness test (ground truth as a perfect prediction should score near-perfect) and a robustness test (malformed inputs must fail cleanly rather than yield spurious scores). Failed checks are graded by severity and trigger minimal targeted repairs. After each repair we immediately re-run the relevant consistency scans and dynamic tests to confirm that the repair itself introduces no new error. This verify--repair cycle iterates over multiple rounds until the verification passes, and what cannot be reliably auto-repaired is escalated to human review.

\paragraph{Environment verification.}
We build the Docker image on a physical machine, run library imports and verify that library versions match our presets. When a build fails, we separate root causes from cascading symptoms and classify each root cause by type. Repair follows one core principle: never override a base-image package. Working from least to most disruptive, we (i) switch to a base-compatible version, (ii) add the missing dependency or runtime configuration, (iii) substitute a compatible alternative and rewrite the affected code, or (iv) remove non-essential packages. A task-critical dependency that resists all of the above triggers a standalone Dockerfile that does not inherit the shared base. Throughout, evaluator and solver dependencies are treated as mandatory and domain convenience packages are best-effort. This verify--diagnose--repair cycle repeats until all checks pass.

\section{Benchmark Quality Calibration Details}
\label{app:calibration-details}

This appendix expands the benchmark quality calibration summarized in \S\ref{sec:nb:calib}.

\paragraph{First-round diagnosis categories.}
The exposed defects fall into six categories: (1)~ground-truth leakage, where the test input carries an unintended channel that allows the agent to recover the answer; (2)~distorted task definitions, where the target degenerates into a deterministic function of input features and can be exactly solved rather than learned; (3)~metrics that fail to distinguish shortcuts from genuine solutions; (4)~evaluator or anchor inconsistencies (e.g., an evaluator metric that disagrees with the task description or metadata); (5)~pipeline or environment errors; and (6)~missing data resources.
Locally verifiable defects receive minimal targeted repairs. Mitigable risks are recorded and backstopped by the information firewall, the web-search-disabled container, and the validity judge. Tasks with broken definitions, unverifiable metrics, or irreparable leakage are dropped. Runs that are legitimate but low-scoring, timed-out, or judged invalid are retained as normal agent failures.

\paragraph{Reproduction-mode audit procedure.}
For each case in reproduce mode, we decompose the paper's method into components (e.g., preprocessing, architecture, loss, training, inference, post-processing), rate each as full, partial, or missing, classify the score outcome, and attribute any anomaly to the agent, the runtime resources, or the package.
Regardless of score, we audit package quality along four axes: (1)~task description and data, (2)~evaluator and scoring, (3)~metadata anchors, and (4)~cross-component consistency. For example, we check whether the SOTA anchor is drawn from the same dataset and granularity as the evaluator computes, whether metadata and evaluator scores share the same scale and units, whether required training data and external resources are present, whether the evaluator returns a reasonable score on ground truth, and whether the task description is consistent with the paper's method.
After human review, 45 tasks are dropped for defects that would systematically contaminate the main evaluation (missing data, evaluator deviations, absent required information, leakage, or distorted scoring), and 17 tasks receive minor repairs (e.g., anchor-value alignment, scale reconciliation, evaluator-logic corrections, incomplete-instance removal, environment and serialization fixes).

\paragraph{Reproducibility analysis.}
On the finalized 90 tasks, we quantify SOTA-anchor attainability (success: $\gimp \ge -0.05$; partial: $-0.2 \le \gimp < -0.05$).
Claude Opus 4.6 reproduces 30/90 tasks successfully and 16/90 partially; DeepSeek-V4-Pro reproduces 21/90 and 13/90. At least one model succeeds on 35/90. Both succeed on 16/90, where $\gimp$ clusters tightly around zero (median $-0.0026$, $90\%$ of absolute deviations $\le 0.031$).
Reproduce-mode success is lower than base mode: Opus drops from $41/90$ to $30/90$ and DeepSeek from $29/90$ to $21/90$, primarily because faithful reproduction triggers heavier training and more complex dependencies. DeepSeek's no-result count rises from $1$ to $29$, accounting for most of its gap. Root-cause attribution of non-success cases is dominated by insufficient compute or time and method simplification rather than package defects.

\section{Case Studies}
\label{app:trajectory-cases}
\begin{figure}[t]
\centering
\includegraphics[width=\linewidth]{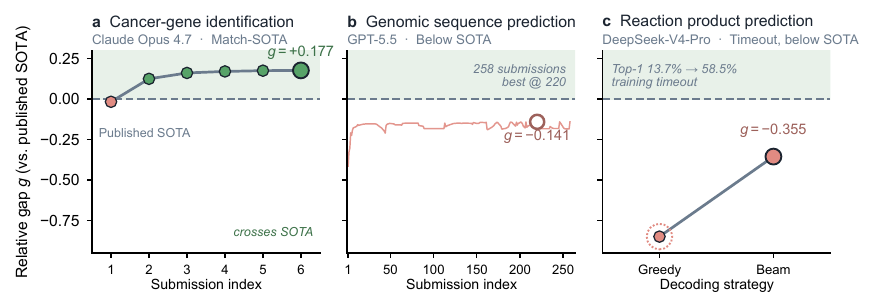}
\caption{\textbf{Representative agent trajectories in NatureBench.}
\textbf{(a)}~Cancer-gene identification (Claude Opus 4.7), six submissions. \textbf{(b)}~Genomic sequence prediction (GPT-5.5), 258 submissions, with the best at attempt~220. \textbf{(c)}~Reaction product prediction (DeepSeek-V4-Pro).}
\label{fig:app-case-trajectories}
\end{figure}

\begin{figure}[t]
\centering
\includegraphics[width=\linewidth]{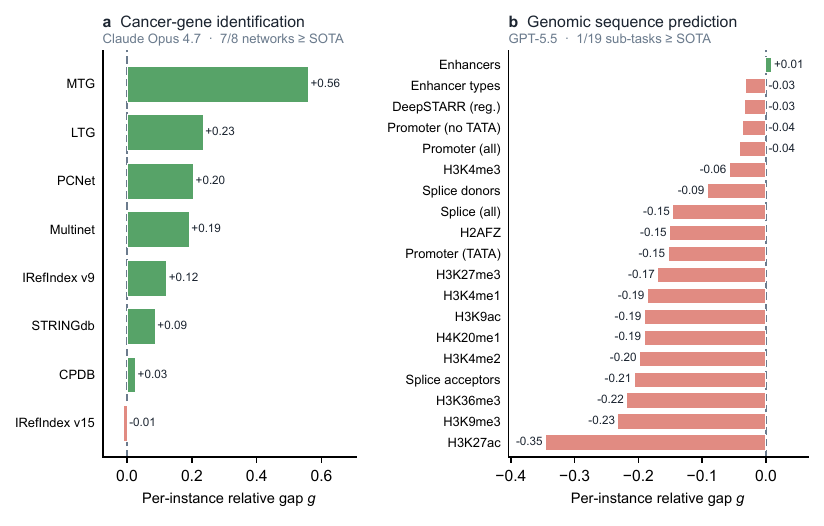}
\caption{\textbf{Per-instance relative gap $g$ at each agent's best submission.}
\textbf{(a)}~Cancer-gene identification (Claude Opus 4.7), eight biological networks. \textbf{(b)}~Genomic sequence prediction (GPT-5.5), 19 genomic sub-tasks. The single-instance reaction task is omitted.}
\label{fig:app-case-instances}
\end{figure}

This appendix complements the aggregate analysis in Section~\ref{sec:exp:discovery-modes} with three representative agent trajectories. The cases cover three recurring outcomes in NatureBench: a method-aligned solution that matches SOTA, a valid but methodologically insufficient solution, and a plausible long-horizon solution limited by execution depth. All cases are drawn from the final 90-task, 10-agent analysis used in Section~\ref{sec:analysis}.
Table~\ref{tab:app-case-overview} lists the selected cases and Table~\ref{tab:app-case-mechanisms} summarizes their trajectory-level mechanisms, while two figures show how each case plays out. Figure~\ref{fig:app-case-trajectories} traces each agent's score across its submission sequence, and Figure~\ref{fig:app-case-instances} decomposes the best submission of the two multi-instance cases into per-instance gaps, showing where the aggregate score comes from.

\begin{center}
\begin{minipage}{\textwidth}
\captionof{table}{Representative trajectory cases analyzed in Appendix~\ref{app:trajectory-cases}.}
\label{tab:app-case-overview}
\centering
\small
\setlength{\tabcolsep}{6pt}
\begin{tabular}{@{}>{\raggedright\arraybackslash}p{0.40\textwidth} l c l@{}}
\toprule
Case & Agent & $g$ & Status \\
\midrule
Cancer gene identification on biological networks & Claude Opus 4.7 & $0.17666$ & Match-SOTA \\
Genomic sequence prediction & GPT-5.5 & $-0.14087$ & Below SOTA \\
Organic reaction product prediction & DeepSeek-V4-Pro & $-0.35540$ & Timeout, below SOTA \\
\bottomrule
\end{tabular}
\end{minipage}
\end{center}

\begin{center}
\begin{minipage}{\textwidth}
\captionof{table}{High-level mechanisms observed in the representative trajectories.}
\label{tab:app-case-mechanisms}
\centering
\small
\setlength{\tabcolsep}{6pt}
\begin{tabular}{@{}>{\raggedright\arraybackslash}p{0.40\textwidth} >{\raggedright\arraybackslash}p{0.24\textwidth} >{\raggedright\arraybackslash}p{0.26\textwidth}@{}}
\toprule
Case & Agent route & Outcome driver \\
\midrule
Cancer gene identification on biological networks & ChebNet/GNN ensemble & Method alignment and training optimization \\
Genomic sequence prediction & From-scratch sequence models & Insufficient representation strength \\
Organic reaction product prediction & Seq2seq reaction modeling & Insufficient execution depth \\
\bottomrule
\end{tabular}
\end{minipage}
\end{center}

\paragraph{Case 1: method-aligned graph modeling can produce a valid success.}
The first task is derived from TREE, a transformer-powered graph representation learning study for identification of cancer-genes \citep{su2025tree}. It asks the agent to identify cancer-associated genes on eight biological networks. Each instance provides a network adjacency matrix, 64-dimensional multi-omics node features, training and validation labels, and a test-node mask. The source problem is naturally a graph-based binary node-classification problem: its core scientific objective is to combine biological network structure with multi-omics node attributes to prioritize cancer genes. The primary metric is AUPRC on each network, aggregated as improvement relative to the paper-side SOTA.

Claude Opus 4.7 selected a route that matched this task structure. The final solution implements a Chebyshev polynomial graph convolutional network (ChebNet) ensemble: it loads the HDF5 network data and node features, computes normalized graph Laplacians, trains with validation AUPRC early stopping, then retrains on the combined train and validation labels before averaging models across Chebyshev orders, depths, and random seeds. The judge marked the submission valid because the predictions were generated by trained graph models, and the raw logs show progressive AUPRC improvements across submissions.
Table~\ref{tab:app-case-tree} summarizes the score progression for this trajectory.

\begin{center}
\begin{minipage}{\textwidth}
\captionof{table}{Score progression and diagnosis for the cancer-gene identification case.}
\label{tab:app-case-tree}
\centering
\small
\begin{tabular}{@{}p{0.24\textwidth}p{0.22\textwidth}p{0.44\textwidth}@{}}
\toprule
Stage & Evidence & Diagnosis \\
\midrule
Initial graph model & $g=-0.01715$ & The first submission was runnable, but still slightly below SOTA. \\
First crossing & $g=0.12457$ & Once the graph-modeling route matured, most networks improved substantially. \\
Ensembling and training optimization & $g=0.161$ to $0.175$ & Chebyshev order, depth, random seeds, and train-plus-validation retraining continued to add gains. \\
Final strengthening & $g=0.17666$ & The last round mainly improved the LTG network and produced the best aggregate score. \\
\bottomrule
\end{tabular}
\end{minipage}
\end{center}

This is a genuine agent success. It correctly treated the task as graph-based node classification and used an appropriate GNN, class-imbalance handling, validation-based early stopping, and ensembling to push the score above SOTA. The per-instance results were also uneven: MTG, LTG, PCNet, and Multinet improved substantially, whereas IRef v15 remained slightly below the paper-side SOTA. However, from another aspect, the agent did not propose a new method of cancer-gene identification.

\paragraph{Case 2: extensive valid iteration can still fall short.}
The second task is derived from the Nucleotide Transformer benchmark for human genomics \citep{dallatorre2025nucleotide}. It contains 19 genomic sequence prediction instances, spanning histone marks, enhancers, promoters, splice sites, and enhancer-activity regression. The source paper's core idea is to learn broad DNA sequence representations from large-scale pretraining and transfer them to diverse downstream sequence-function tasks. The agent must submit predictions for 18 classification tasks and one regression task.

This trajectory is long and technically substantial. The agent produced 258 submissions, with the best score at attempt 220. It began with compact k-mer count models and fast linear classifiers, then added splice-site motif rules, GPU CNNs, enhancer-activity CNN ensembles, a two-stage enhancer-type classifier, and many threshold sweeps. The judge marked the submission valid because all predictions were generated by models trained on the provided data.
Table~\ref{tab:app-case-nucleotide} summarizes the main trajectory stages.

\begin{center}
\begin{minipage}{\textwidth}
\captionof{table}{Score progression and diagnosis for the genomic sequence prediction case.}
\label{tab:app-case-nucleotide}
\centering
\small
\begin{tabular}{@{}p{0.24\textwidth}p{0.22\textwidth}p{0.44\textwidth}@{}}
\toprule
Stage & Evidence & Diagnosis \\
\midrule
Fast baseline & $g=-0.41445$ & The agent first solved the submission-completeness problem. \\
Task specialization & $g=-0.20882$ & Local biological sequence cues improved several sub-tasks. \\
Deep iteration & $g=-0.14087$ & Iteration was effective but gradually saturated. \\
Remaining gap & No Match-SOTA & From-scratch models lacked the representation strength of the paper-side route. \\
\bottomrule
\end{tabular}
\end{minipage}
\end{center}

The failure is therefore not a formatting or execution failure. It is a method-layer limitation: the agent built a sophisticated runnable pipeline, but its chosen models lacked the inductive bias and representation capacity of large-scale genomic pretraining. This case illustrates why many agent failures on NatureBench tasks are better described as ``runnable but not strong enough'' than as simple coding failures.

\paragraph{Case 3: a plausible route can be limited by execution depth.}
The third task is derived from LocalTransform, a generalized-template-based graph neural network for organic reactivity prediction \citep{chen2022localtransform}. It asks the agent to predict major organic reaction products for USPTO-480k atom-mapped reactants. The source paper's core idea is to model local reaction centers and bond changes with reaction templates, molecular graph representations, and chemistry tooling. This route reaches a Top-1 exact-match accuracy of $0.908$, while a strong sequence-to-sequence baseline reaches $0.887$. The task requires both learning reaction transformations from hundreds of thousands of examples and generating ranked product SMILES efficiently.

The agent selected a plausible but expensive route: it implemented a complete sequence-to-sequence reaction model with a SMILES tokenizer, dataset loader, Transformer model, training loop, checkpointing, and prediction pipeline. The judge marked the submission valid because the final predictions were generated by the trained model with checkpointing and beam-search inference.
Table~\ref{tab:app-case-localtransform} summarizes the trajectory stages and score progression.

\begin{center}
\begin{minipage}{\textwidth}
\captionof{table}{Score progression and diagnosis for the reaction product prediction case.}
\label{tab:app-case-localtransform}
\centering
\small
\begin{tabular}{@{}p{0.24\textwidth}p{0.22\textwidth}p{0.44\textwidth}@{}}
\toprule
Stage & Evidence & Diagnosis \\
\midrule
Route selection & 24.3M parameters & The route was plausible, but computationally heavy. \\
Long training & Loss $2.0312 \rightarrow 1.3149$ & The model learned, but training consumed much of the budget. \\
Greedy decoding & Top-1 $13.68\%$ & The first valid submission used a weak decoding strategy. \\
Beam search & \begin{tabular}[t]{@{}l@{}}Top-1 $58.53\%$ \\ $g=-0.35540$\end{tabular} & Inference engineering helped sharply, but the final score remained below SOTA. \\
\bottomrule
\end{tabular}
\end{minipage}
\end{center}

The key limitation was execution depth rather than invalidity. The agent found a reasonable scientific-computational route, but the task required deeper training, more efficient generation, and more specialized chemical modeling than the fixed budget allowed. This case illustrates the execution-layer failures discussed in Section~\ref{sec:exp:discovery-modes}: some agents identify a plausible direction, yet fail because the required training and inference loop is too long.

\section{Resource Usage Details}
\label{app:cost}

Following common practice in agent benchmarks, we report resource usage at the trajectory level rather than only at the initial prompt level. For each evaluated agent, we aggregate token-usage information from valid execution logs and summarize per-case input tokens, output tokens, and estimated API cost. Input tokens use exact harness- or provider-reported usage fields. When the log records cache accounting, we retain the distinction among non-cached input, cache read or hit tokens, and cache creation or write tokens; the mean input-token column reports their sum so that it reflects the full amount of context processed during the run.

Output-token accounting follows the most reliable source available for each model. For Claude Opus, GPT, and Gemini runs, we use exact provider- or harness-reported output-token fields. For third-party models executed through Claude Code, the logged output-token fields are incomplete, so we estimate output tokens from agent-authored trajectory text using the standard rule of thumb that one token corresponds to roughly four English characters~\citep{openai2026tokens,anthropic2026pricing}. Rows that use this output-token estimate, and the costs derived from it, are marked with an asterisk. Cost is computed with official standard list prices and provider-specific cache rates. We exclude limited-time promotions, batch/flex/priority modes, regional or data-residency uplifts, cache-storage charges that cannot be recovered from the logs, and OpenAI long-context multipliers whose per-request triggers cannot be recovered from aggregate Codex logs.
Table~\ref{tab:cost} reports the resulting per-agent means over valid runs.

\begin{center}
\begin{minipage}{\textwidth}
\captionof{table}{Per-agent token usage and estimated API cost. Means are computed over valid runs with recorded usage information. \textsuperscript{*}Values estimated from agent-authored trajectory text; costs with \textsuperscript{*} are derived from these estimates.}
\label{tab:cost}
\centering
\small
\begin{tabular}{lccc}
\toprule
\textbf{Agent} & \textbf{Mean input tokens} & \textbf{Mean output tokens} & \textbf{Mean cost (USD)} \\
\midrule
Claude Opus 4.7          & $24.25$M & $179.3$K & \$21.65 \\
Claude Opus 4.6          & $9.03$M  & $87.6$K  & \$16.56 \\
GPT-5.5                 & $8.25$M  & $31.9$K  & \$6.01 \\
GPT-5.4                 & $10.79$M & $43.6$K  & \$4.14 \\
Gemini 3.5 Flash        & $11.05$M & $34.2$K  & \$4.49 \\
Qwen 3.7 Max            & $3.85$M  & $88.7$K\textsuperscript{*} & \$10.19\textsuperscript{*} \\
Kimi K2.6               & $13.32$M & $85.0$K\textsuperscript{*} & \$12.99\textsuperscript{*} \\
MiniMax-M2.7            & $4.44$M  & $55.7$K\textsuperscript{*} & \$1.35\textsuperscript{*} \\
DeepSeek-V4-Pro         & $11.34$M & $77.5$K\textsuperscript{*} & \$0.15\textsuperscript{*} \\
GLM-5.1                 & $2.77$M  & $65.8$K\textsuperscript{*} & \$4.12\textsuperscript{*} \\
\bottomrule
\end{tabular}
\end{minipage}
\end{center}

\end{document}